%% file: root.tex
\documentclass[letterpaper, 10 pt, journal, twoside]{ieeetran}
\IEEEoverridecommandlockouts
\usepackage{amsmath,amsfonts}
\usepackage{algpseudocode}
\usepackage[linesnumbered, ruled, vlined]{algorithm2e}
\usepackage{array}
\usepackage[caption=false,font=normalsize,labelfont=sf,textfont=sf]{subfig}
\usepackage{textcomp}
\usepackage{stfloats}
\usepackage{url}
\usepackage{verbatim}
\usepackage{graphicx}
\usepackage{subcaption}
\usepackage{subcaption}
\usepackage[hidelinks]{hyperref}
\usepackage{cite}
\usepackage{enumitem}
\usepackage{booktabs}
\usepackage{listings}
\usepackage{xcolor}
\usepackage{multirow}
\input{math_commands}
\usepackage{cleveref}
\Crefname{definition}{Definition}{Definitions}
\Crefname{proposition}{Proposition}{Propositions}
\Crefname{theorem}{Theorem}{Theorems}
\Crefname{figure}{Fig.}{Figs.}
\Crefname{equation}{Eq.}{Eqs.}
\Crefname{section}{Section}{Sections}
\Crefname{subsection}{Section}{Sections}
\Crefname{subsubsection}{Section}{Sections}
\Crefname{appendix}{Appendix}{Appendices}
\Crefname{algorithm}{Algorithm}{Algorithms}

\algrenewcommand{\algorithmiccomment}[1]{\textit{/* #1 */}}

\definecolor{codegray}{rgb}{1,0.99,1}
\definecolor{codecomment}{rgb}{0,0.5,0}
\definecolor{keywordcolor}{rgb}{0.25,0.2,0.7}

\lstdefinestyle{mystyle}{
    backgroundcolor=\color{codegray},
    commentstyle=\color{codecomment}\itshape,
    keywordstyle=\color{keywordcolor}\bfseries,
    numberstyle=\tiny\color{gray},
    stringstyle=\color{purple},
    basicstyle=\ttfamily\footnotesize,
    breaklines=true,
    captionpos=b,
    keepspaces=true,
    numbers=left,
    numbersep=5pt,
    showspaces=false,
    showstringspaces=false,
    showtabs=false,
    frame=single,
    xleftmargin=1em,
    xrightmargin=1em,
    aboveskip=1em,
    belowskip=1em,
    language=Python,
}

\usepackage{pifont} 
\newcommand{\yes}{\ding{51}} 
\newcommand{\no}{\ding{55}}  
\newcommand{\orcid}[1]{\href{https://orcid.org/#1}{\includegraphics[width=0.6em]{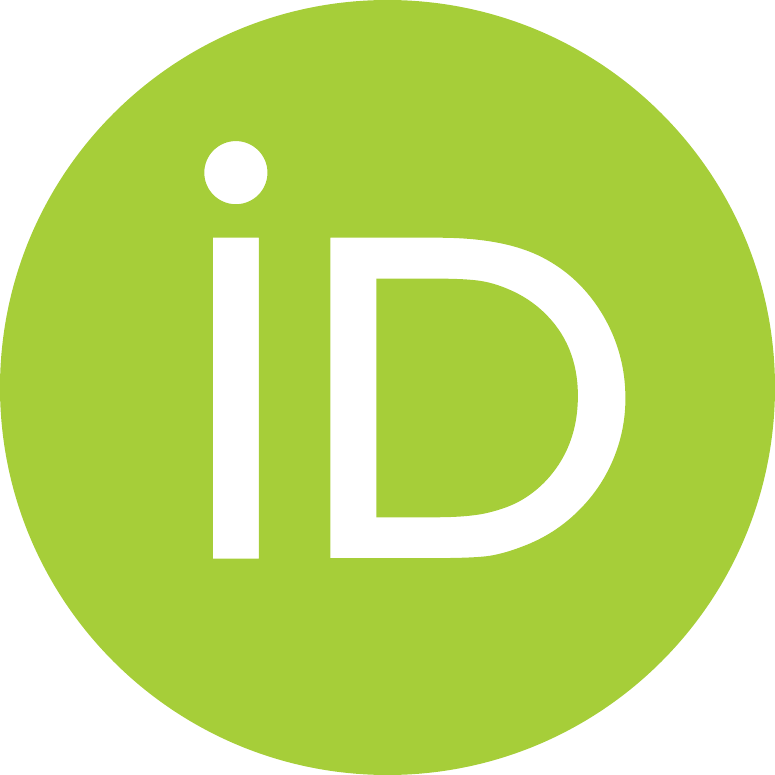}}}

\usepackage{glossaries}
\newacronym{qp}{QP}{quadratic programming}
\newacronym{pmm}{PMM}{proximal method of multipliers}
\newacronym{ddp}{DDP}{differential dynamic programming}
\newacronym{mpc}{MPC}{model predictive control}
\newacronym{sqp}{SQP}{sequential quadratic programming}
\newacronym{miqp}{MIQP}{mixed-integer quadratic programming}
\newacronym{ipm}{IPM}{interior-point method}
\newacronym{alm}{ALM}{augmented Lagrangian method}
\newacronym{ip}{IP}{interior-point}
\newacronym{nip}{NIP}{non-interior-point}
\newacronym{nipm}{NIPM}{non-interior-point method}
\newacronym{ncp}{NCP}{nonlinear complementarity problem}
\newacronym{admm}{ADMM}{alternating direction method of multipliers}
\newacronym{slam}{SLAM}{simultaneous localization and mapping}
\newacronym{lp}{LP}{linear program}
\newacronym{socp}{SOCP}{second-order cone program}
\newacronym{lcp}{LCP}{linear complementarity problem}
\newacronym{kkt}{KKT}{Karush--Kuhn--Tucker}
\newacronym{fonc}{FONC}{first-order necessary conditions}
\newacronym{ift}{IFT}{implicit function theorem}
\newacronym{licq}{LICQ}{linear independence constraint qualification}

\newif\ifpublished
\newcommand{\published}[1]{%
  \ifnum\pdfstrcmp{#1}{True}=0\relax
    \publishedtrue
  \else
    \publishedfalse
  \fi
}
\published{True} 

\begin{document}

\title{ODYN: An All-Shifted Non-Interior-Point Method\\ for Quadratic Programming in Robotics and AI}
\ifpublished
\author{
    Jose Rojas$^\dagger$\orcid{0000-0001-5580-6616}\quad
    Aristotelis Papatheodorou\orcid{0000-0003-0290-7071}\quad
    Sergi Martinez$^\dagger$\orcid{0009-0001-0027-8358}\quad
    Andrea Patrizi$^*$\orcid{0000-0002-5571-3716}\quad\\
    Ioannis Havoutis\orcid{0000-0002-4371-4623}\quad
	Carlos Mastalli$^\dagger$\orcid{0000-0002-0725-4279}
\thanks{This research was conducted as part of the Advancing MANipulation skills in Legged Robots (AMAN) project, a collaborative project supported by Tata Consultancy Services.
\textit{(Corresponding author: Carlos Mastalli)}}
\thanks{
Jose Rojas, Sergi Martinez and Carlos Mastalli are part of the Robot Motor Intelligence (RoMI) Lab, Heriot-Watt University, U.K.
}
\thanks{
Aristotelis Papatheodorou and Ioannis Havoutis are part of the Oxford Robotics Institute (ORI), University of Oxford, U.K.
}
\thanks{
Andrea Patrizi is part of Humanoids and Human Centered Mechatronics, Italian Institute of Technology, Italy.
}
}
\else
\author{Author Names Omitted for Anonymous Review.}
\fi

\ifpublished
\markboth
{Rojas \MakeLowercase{\textit{et al.}}: An All-Shifted Non-Interior-Point Method for Quadratic Programming in Robotics and AI}
{}
\else
\markboth
{Submitted to IEEE Transactions on Robotics.}
{Anonymous \MakeLowercase{\textit{et al.}}: An All-Shifted Non-Interior-Point Method for Quadratic Programming in Robotics and AI}
{}
\fi


\maketitle

\begin{abstract}
We introduce \textsc{Odyn}, a novel all-shifted primal--dual non-interior-point quadratic programming (QP) solver designed to efficiently handle challenging dense and sparse QPs.
\textsc{Odyn} combines all-shifted nonlinear complementarity problem (NCP) functions with the proximal method of multipliers to robustly address ill-conditioned and degenerate problems, without requiring linear independence of the constraints. 
It exhibits strong warm-start performance and is well-suited to both general-purpose optimization and robotics and AI applications, including model-based control, estimation, and kernel-based learning methods. 
We provide an open-source implementation and benchmark \textsc{Odyn} on the Maros--Mészáros test set, demonstrating state-of-the-art convergence performance in small-to-high-scale problems.
The results highlight \textsc{Odyn}'s superior warm-starting capabilities, which are critical in sequential and real-time settings common in robotics and AI. 
These advantages are further demonstrated by deploying \textsc{Odyn} as the backend of an SQP-based predictive control framework (\texttt{OdynSQP}), as the implicitly differentiable optimization layer for deep learning (\texttt{ODYNLayer}), and the optimizer of a contact-dynamics simulation (\texttt{ODYNSim}).
\end{abstract}

\begin{IEEEkeywords}
quadratic programming, non-interior-point methods, NCP functions, differentiable optimization, sequential quadratic programming
\end{IEEEkeywords}

\section{Introduction}
\IEEEPARstart{Q}{uadratic} programming is a foundational optimization framework in robotics, AI, and a broad range of scientific and engineering disciplines, including operations research.
In robotics,~\gls{qp} formulations commonly arise in applications such as contact simulation,~\gls{mpc}, whole-body control, state-estimation, and~\gls{slam}.
Beyond these direct uses,~\gls{qp} solvers also underpin more advanced optimization paradigms, serving as the building blocks of \gls{sqp} methods for nonlinear programs and \gls{miqp} approaches for problems with discrete decision variables, as detailed in \cite{nocedal-optbook} and \cite{lee-minpbook}, respectively.

Modern numerical \gls{qp} solvers can be broadly categorized into two main classes: active-set methods and relaxation-based methods.
Active-set methods explicitly maintain and update a working set of active constraints, making them well suited to warm-started and small-to-medium-scale problems in which the active set evolves smoothly.
Relaxation-based methods, including \glspl{ipm} and \glspl{alm}, instead enforce feasibility through barrier or penalty mechanisms, respectively, and are widely adopted for their robustness and scalability.
The standard formulation supported by many off-the-shelf \gls{qp} solvers is defined as follows:
\begin{equation}\label{eq:original_qp}
\begin{aligned}
\min\limits_{\vx \in \mathbb{R}^n} \quad
    & \tfrac{1}{2}\,\vx^\transpose \matQ\,\vx + \vecC^\transpose \vx \\
\text{subject to} \quad
    & \matA\,\vx = \vecB, \\
    & \matG\,\vx \le \vecH,
\end{aligned}
\end{equation}
where $\vx\in\R^n$ denotes the decision vector, $\matQ\in\R^{n\times n}$ is a symmetric and positive semidefinite quadratic cost matrix ($\matQ\succeq\mathbf{0}$), $\vecC\in\R^n$ is a vector of linear cost terms, $\matA\in\R^{m\times n}$ is the matrix of equality constraints, $\vecB\in\R^m$ is the right-hand side vector for the equality constraints,
$\matG\in\R^{p\times n}$ is the matrix of inequality constraints, and $\vecH\in\R^p$ is the right-hand side vector for the inequality constraints.
For large-scale problems (e.g., $n \gg 100$), \gls{qp} solvers typically leverage sparse backends, relying on sparse linear solvers~\cite{golub-matcompbook} and representing $\matQ$, $\matA$, and $\matG$ as sparse matrices.
In contrast, for small to moderately sized problems---frequently encountered in robotics and AI---dense linear solvers and matrix representations tend to be more efficient and convenient.

Robotics and AI applications often demand optimizers that support warm-starting, high efficiency, robustness, and scalability.
For example, \gls{mpc} algorithms continuously update motion plans and control actions by solving large, nonlinear optimization problems, typically warm-started from previous solutions.
Similarly, warm-start strategies are critical for real-time localization and system identification~\cite{martinez2025multi, martinez2026sysid}.
Moreover, contact simulators rely on optimizers that must handle rank-deficient contact Jacobians or positive semidefinite Delassus matrices~\cite{tsounis2025}, while still benefiting from warm-start information.
Yet state-of-the-art solvers such as \textsc{Gurobi}~\cite{gurobi}, \textsc{Mosek}~\cite{mosek}, \textsc{PiQP}~\cite{schwan2023_piqp}, \textsc{ProxQP}~\cite{bambade2022_proxqp}, and \textsc{OsQP}~\cite{stellato2020_osqp} often struggle to satisfy all of these requirements simultaneously.

A promising class of algorithms, known as \glspl{nipm}, offers an alternative to classical interior-point methods.
They are motivated in part by the desire to avoid the strict interior-feasibility requirements imposed by its log-barrier functions.
Although these methods scale well to problems with many inequality constraints, they often make robust warm-starting difficult when solving sequences of closely related optimization problems.
\textsc{Odyn} is built upon these principles and demonstrates strong capabilities that advance the state of the art in quadratic programming.

\begin{figure*}[!t]
    \centering
    \includegraphics[width=\textwidth]{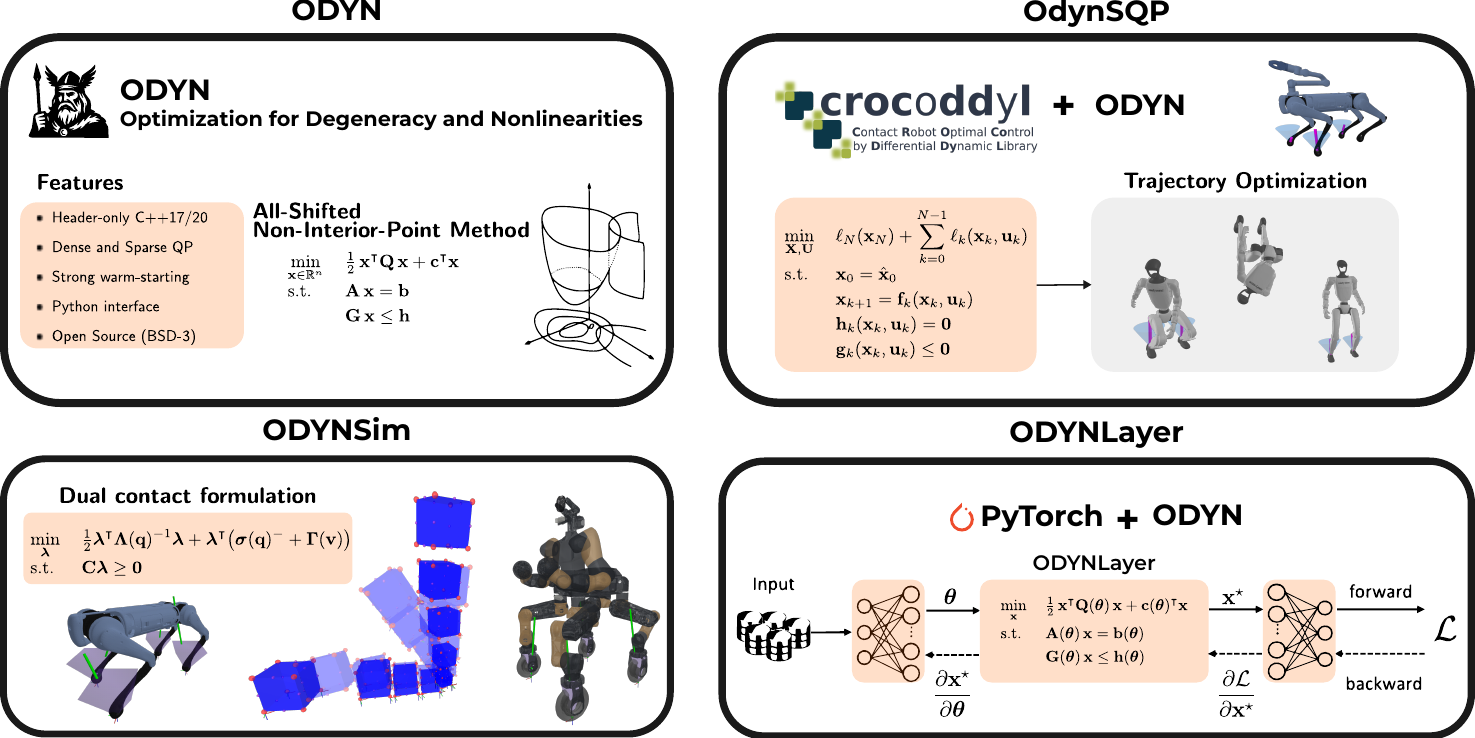}
    \caption{Overview of \textsc{Odyn} applications in robotics and AI.
    \textsc{Odyn} serves as the computational core for constrained nonlinear trajectory optimization (\texttt{OdynSQP}), contact-dynamics simulation (\texttt{ODYNSim}), and differentiable optimization layers (\texttt{ODYNLayer}), providing a common optimization backbone across control, simulation, and learning.}
    \label{fig:cover_image}
\end{figure*}

\subsection{Contribution}
We introduce \textsc{Odyn}, a warm-startable~\gls{qp} solver that we characterize as an all-shifted \glspl{nipm}---a class of algorithms that remains relatively underexplored in the numerical optimization community, yet holds significant potential for robotics and AI workloads.
Our main contributions are:
\begin{enumerate}[label=(\roman*)]
    \item \textbf{A new NIPM-based QP formulation.}
    We develop a novel solver that blends elements of primal--dual interior-point methods with augmented-Lagrangian techniques.
    \textsc{Odyn} incorporates three key novelties: a centering-weighted barrier, an all-shifted~\gls{ncp} formulation, and proximal Lagrangian penalties for equality and inequality constraints.
    They are critical for robust performance under rank deficiencies, degeneracies, and poor scaling.

    \item \textbf{A complete open-source implementation.}
    We release an efficient \textsc{C++} implementation with \textsc{Python} bindings.\footnote{The code will be available after acceptance.}
    \textsc{Odyn} currently supports dense and sparse backends, multiple floating-point numbers, code generation, and nullspace handling of equality constraints.
    Moreover, we incorporate \texttt{ODYNLayer}, a framework for training~\gls{qp} layers within neural-network architectures in \textsc{Pytorch}.
    Finally, \texttt{OdynSQP} is developed on top of \textsc{Crocoddyl}, an open-source software framework for model predictive control in robotics.

    \item \textbf{Benchmarking and robotics/AI evaluation.}
    We benchmark \textsc{Odyn} against state-of-the-art interior-point and augmented-Lagrangian \gls{qp} solvers using the Maros--Mészáros test set.
    We further assess its practical utility across robotics and AI use cases:
    (i) an efficient \gls{sqp} framework for real-time \gls{mpc} named \texttt{OdynSQP},
    (ii) contact simulation solved via \texttt{ODYNSim}, and
    (iii) training differentiable \gls{qp} layers through the \texttt{ODYNLayer}.
\end{enumerate}
These characteristics make \textsc{Odyn} particularly well suited for robotics and AI applications (\Cref{fig:cover_image}). 
More broadly, its strong convergence properties also make it relevant to a wide range of engineering and scientific problems.

Below, we review the main classes of available \gls{qp} solvers, the underlying theory, and their salient characteristics.

\section{Related Work}
\Acrfull{qp} plays a central role in robotics, control, and numerical optimization, motivating the development of numerous solvers built on distinct algorithmic paradigms.
Modern \gls{qp} methods can be broadly grouped into three families: active-set methods, \acrlongpl{alm} (penalty-based), and \acrlongpl{ipm} (barrier-based).
Each family offers different trade-offs in scalability, warm-start capability, and robustness to ill-conditioning. 
\glspl{alm}, for example, are traditionally designed for equality-constrained problems, although solvers such as \textsc{Algencan}~\cite{andreani2008_algencan} and \textsc{Lancelot}~\cite{conn2010_lancelot} have introduced mechanisms to handle inequality constraints.
Their appeal lies in their ability to cope with rank deficiencies in equality constraints while mitigating ill-conditioning. 
In contrast, \glspl{ipm} excel at handling large numbers of inequality constraints through barrier functions and typically incorporate equality constraints via primal--dual~\gls{kkt} systems, as in \textsc{Ipopt}~\cite{wachter-mp06}.
By avoiding the combinatorial complexity of explicitly tracking active constraints, they offer better scalability than active-set or penalty-based methods.
However, their reliance on strict interior feasibility makes warm-starting inherently difficult.
\Cref{warm-starting-ipm} reviews recent efforts to mitigate this limitation through mechanisms such as exact-penalty formulations and all-shifted complementarity strategies.

Motivated by the warm-starting limitations of~\glspl{ipm},~\acrfullpl{nipm} have gained renewed attention. These approaches relax the strict feasibility requirements of traditional \glspl{ipm} by reformulating complementarity constraints using \gls{ncp} functions.
\Cref{nip-relatedwork} surveys the development of~\glspl{nipm}, including nonsmooth and path-following variants, and highlights recent hybrid strategies that blend \glspl{ipm} with \gls{alm}-style updates. 
We begin by reviewing the most prominent \gls{qp} solvers in the literature.

\subsection{Available \gls{qp} solvers and their features}\label{sec:available-qp}
\textsc{qpOASES} is a classical example of an active-set solver, introduced in~\cite{ferreau2014_qpoases}.
It is tailored for online and real-time applications, such as \gls{mpc}, and is particularly effective for small- to medium-sized dense \glspl{qp}.
Designed with warm-starting in mind, \textsc{qpOASES} performs well when the problem structure evolves gradually over time.

Another widely used active-set solver in robotics is \textsc{eiquadprog}~\cite{eiquadprog}, which implements the Goldfarb--Idnani dual active-set algorithm~\cite{goldfarb1983dual}.
Originally developed within the \textsc{Stack-of-Tasks} framework, \textsc{eiquadprog} is commonly employed in whole-body control and legged locomotion.
Like \textsc{qpOASES}, it is most effective for dense problems and benefits from warm-starting.

However, active-set solvers such as \textsc{qpOASES} and \textsc{eiquadprog} tend to struggle with large-scale or highly degenerate problems.
They may require a large number of iterations when the active set changes substantially between successive iterates.
This behavior stems from the combinatorial complexity of identifying the optimal set of active constraints.

\textsc{OsQP} belongs to the family of operator-splitting methods and leverages the \gls{admm} algorithm~\cite{stellato2020_osqp}.
This places \textsc{OsQP} within the class of \glspl{alm}.
It is known for its robustness and scalability to large, sparse problems and supports both structure-exploiting updates and warm-starting.
\textsc{OsQP} also offers code generation capabilities for maximal runtime efficiency in CPUs~\cite{osqp2017-codegen}, and can be executed on GPUs~\cite{osqp2020-gpu}.
However, as a first-order method, its key limitation lies in the fact that typically achieves only moderate solution accuracy.
It may also struggle with problems that exhibit strong ill-conditioning or involve tight constraints.

\textsc{ProxQP} is another \gls{alm}-based solver, with notable distinctions inspired from \textsc{Lancelot} as highlighted in~\cite{bambade2022_proxqp}.
It employs a proximal augmented Lagrangian method and incorporates second-order information, enabling more accurate and robust solutions compared to first-order methods such as \textsc{OsQP}.
\textsc{ProxQP} is designed with a modular architecture and offers a differentiable~\gls{qp} layer for integration into deep learning pipelines~\cite{bambade2024qplayer}.
It supports warm-starting and performs well for medium-scale problems.
However, its has limited robustness to solve severe ill-conditioned problems and does not yet provide strong guarantees for exact infeasibility certification.

\textsc{PiQP} was recently introduced in~\cite{schwan2023_piqp}.
It adopts a hybrid approach by combining an infeasible primal--dual~\gls{ipm} with the proximal method of multipliers.
This combination enhances numerical robustness and enables \textsc{PiQP} to handle ill-conditioned convex \gls{qp} problems without requiring the linear independence of the constraints---an advantage over traditional~\glspl{ipm}.
While this hybridization results in a more robust solver, \textsc{PiQP} inherits the typical limitations of~\glspl{ipm} in lacking efficient warm-starting and may require repeated factorization across iterations.
Nonetheless, \textsc{PiQP} demonstrates excellent performance on challenging \gls{qp} problems, particularly when benchmark it against the Maros--Mészáros test set.
The Maros--Mészáros benchmark comprises both separable and non-separable \gls{qp} problems~\cite{maros1999repository}.
Separable problems are characterized by a diagonal $\matQ$ matrix, containing only squared terms, while non-separable problems include off-diagonal entries, with $\matQ$ remaining symmetric.

In addition to the aforementioned \gls{qp} solvers, several other notable approaches merit consideration.
\textsc{qpSWIFT} is an \gls{ipm}-based solver tailored for embedded \glspl{qp} presented in~\cite{pandala2019_qpswift}.
Commercial solvers such as \textsc{Gurobi} and \textsc{Mosek} provide highly optimized interior-point algorithms and support a broad class of problems, though they are closed-source and not specifically designed for real-time robotics or AI.
\textsc{Hpipm} offers a dense, Riccati-based formulation particularly suited for \gls{mpc} applications~\cite{frison2020_hpipm}.
However, as an \gls{ipm}-type method, it inherits the typical limitations related to warm-starting.

\subsection{Warm-starting \glspl{ipm}}\label{warm-starting-ipm}
\Gls{ip} methods face inherent challenges with warm-starting, as their iterates must remain strictly within the \emph{inequality-feasible set} (or the positive orthant when slack variables are introduced). 
A key difficulty is that reusing the optimal solution of a previous problem (e.g., in~\gls{sqp} solvers) as an initial point for a related problem often leads to blocked search directions and severe ill-conditioning, making it hard for the solver to identify changes in the active set~\cite{gondzio2008}. 
A common workaround is to retain one or several unconverged iterates and modify them to satisfy the feasibility requirements of the new problem. 
Another line of research includes the exact-penalty reformulation of~\cite{benson2007}, although its reliance on adaptive penalty tuning can introduce numerical instabilities. 
However, a related approach introduced in~\cite{engau2009primal} avoids penalty parameters but still requires explicit identification of the active constraints. 
Alternatively, the approach of~\cite{skajaa2013} uses a convex combination of a well-centered reference point and the previous solution to warm-start \glspl{lp} and \glspl{socp} within the homogeneous self-dual framework.
Despite these efforts, progress on robust and general warm-starting mechanisms for~\glspl{ipm} has remained limited.

More recently, the method of~\cite{gill2024} leverages \emph{all-shifted complementarity constraints} to relax non-negativity requirements through a projected line search mechanism.
This strategy offers a promising alternative to classical warm-starting techniques by allowing iterates to move outside the strict interior region while still preserving the structure required by primal--dual interior-point algorithms.

\subsection{Non-interior-point methods}\label{nip-relatedwork}
\Gls{nip} methods were developed as an alternative to \glspl{ipm}, with the goal of solving constrained optimization problems, including linear and nonlinear complementarity problems, by leveraging~\gls{ncp} functions.
\gls{ncp} functions avoid the need for iterates to remain strictly interior to the feasible set.

\gls{nip} methods can be broadly classified into two main categories: semismooth and path-following methods.
Semismooth~\gls{nip} algorithms utilize semismooth~\gls{ncp} functions, such as the minimum map and the Fischer--Burmeister function~\cite{qi1993nonsmooth}.
These approaches have been successfully applied in diverse domains including physics simulation~\cite{todorov2010implicit,macklin2019}, quadratic programming~\cite{liao2020fbstab}, nonlinear complementarity problems~\cite{sun1999regularization}, and second-order cone complementarity problems~\cite{ma2018discovery}.

In contrast, path-following~\gls{nip} algorithms solve a sequence of nonlinear systems defined by smoothed~\gls{ncp} functions, parameterized by a smoothing parameter.
The trajectory of optimal solutions for these systems is known as the central path.
These algorithms have been explored for solving the \gls{lcp}~\cite{kanzow1996, burke1998global}, optimal control problems with complementarity constraints~\cite{lin2025non}, and linear programming using predictor-corrector strategies~\cite{burke1998non, engelke2002predictor}.
More recently, the relaxed interior-point method has been introduced, which combines interior-point method barrier terms with augmented Lagrangian penalties.
This leads to a path-following~\gls{nip} algorithm that incorporates a smoothed minimum~\gls{ncp} function, and has been applied to quadratic programming~\cite{zhang2023iprqp} and semidefinite programming~\cite{zhang2024iprsdp}.

For context, we provide below a summary of representative state-of-the-art \gls{qp} solvers discussed above.
\Cref{tab:qp_solver_comparison} categorizes these solvers according to their underlying algorithmic principles
and highlights their main practical features, including backend support, warm-start capabilities,
robustness to degeneracy, achievable solution accuracy, and suitability for differentiation.
\textsc{Odyn} belongs to the class of path-following methods.

Before detailing the \textsc{Odyn}'s \gls{nip} algorithm, we first introduce the mathematical foundations underlying the construction of \textsc{Odyn}, some of which are novel.

\begin{table*}[t]
\caption{Comparison of representative~\gls{qp} solvers grouped by algorithmic method.}
\label{tab:qp_solver_comparison}
\centering
\renewcommand{\arraystretch}{1.15}
\setlength{\tabcolsep}{6pt}
\begin{tabular}{p{2.1cm}|r|cccccc}
\toprule
\multicolumn{1}{c|}{\textbf{Method}} 
& \multicolumn{1}{c|}{\textbf{Solver}} 
& \multicolumn{1}{c}{\textbf{Algorithm}} 
& \multicolumn{1}{c}{\textbf{Backend}} 
& \multicolumn{1}{c}{\textbf{Warm-start}} 
& \multicolumn{1}{c}{\textbf{Degeneracy}} 
& \multicolumn{1}{c}{\textbf{Accuracy}} 
& \multicolumn{1}{c}{\textbf{Differentiable}} \\
\midrule

\multirow{2}{*}{\parbox[c]{2.1cm}{\raggedleft \textbf{Non-interior}\\ \textbf{Point}}}
& \textsc{Odyn} 
& All-shifted NIPM 
& Dense / Sparse 
& \yes 
& \yes 
& High 
& \yes \\

& \textsc{Fbstab}~\cite{liao2020fbstab}
& Semismooth NIPM 
& Sparse 
& \yes 
& \yes 
& High 
& \no \\
\midrule

\multirow{4}{*}{\parbox[c]{2.1cm}{\raggedleft \textbf{Interior-Point}}}
& \textsc{PiQP}~\cite{schwan2023_piqp} 
& Proximal IPM 
& Sparse 
& \no 
& \yes 
& High 
& \no \\

& \textsc{Mosek}~\cite{mosek}
& Homogeneous self-dual IPM 
& Dense / Sparse 
& \no 
& \yes 
& High 
& \no \\

& \textsc{Gurobi}~\cite{gurobi}
& Primal--dual IPM 
& Dense / Sparse 
& \no 
& \yes 
& High 
& \no \\

& \textsc{qpSWIFT}~\cite{pandala2019_qpswift}
& Primal--dual IPM 
& Sparse 
& \no 
& \yes 
& High 
& \no \\
\midrule

\multirow{2}{*}{\parbox[c]{2.1cm}{\raggedleft \textbf{Augmented}\\ \textbf{Lagrangian}}}
& \textsc{ProxQP}~\cite{bambade2022_proxqp}
& Proximal ALM 
& Dense / Sparse 
& \yes 
& Partial 
& High 
& \yes \\

& \textsc{OsQP}~\cite{stellato2020_osqp}
& ADMM-based ALM 
& Sparse 
& \yes 
& Limited 
& Medium 
& \no \\
\midrule

\multirow{2}{*}{\parbox[c]{2.1cm}{\raggedleft \textbf{Active-set}}}
& \textsc{qpOASES}~\cite{ferreau2014_qpoases}
& Online active-set 
& Dense 
& \yes 
& Limited 
& High 
& \no \\

& \textsc{eiquadprog}~\cite{eiquadprog}
& Goldfarb--Idnani active-set 
& Dense 
& \yes 
& Limited 
& High 
& \no \\
\bottomrule
\end{tabular}
\end{table*}

\section{Algorithmic Foundations}
We begin by deriving the optimality conditions and establishing the connections that enable us to construct a non-interior approach from interior-point principles.
We then discuss mechanisms for introducing regularity in degenerate optimization problems.
Both components draw inspiration from the literature on interior-point and augmented-Lagrangian methods.

\subsection{Overview of numerical challenges}
We start by simplifying inequality constraints complexity of~\Cref{eq:original_qp} as follows
\begin{equation}\label{eq:slack_reformulated_qp}
\begin{aligned}
\min\limits_{\vx, \vs} \quad
    & \tfrac{1}{2}\,\vx^\transpose \matQ\,\vx + \vecC^\transpose \vx \\\text{subject to} \quad
    & \matA\,\vx = \vecB, \\
    & \matG\,\vx + \vs = \vecH, \\
    & \vs \ge \vzero.
\end{aligned}
\end{equation}
where $\vs\in\R^p$ is a vector of slack variables.
To establish the optimality conditions, we define the Lagrangian of \Cref{eq:slack_reformulated_qp} as
\begin{align}
\mathcal{L}(\vx, \vs, \vy, \vz)
&= \tfrac12 \vx^\transpose \matQ \vx + \vecC^\transpose \vx 
   + \vy^\transpose (\matA \vx - \vecB) \notag\\
&\quad + \vz^\transpose (\matG \vx + \vs - \vecH),
\end{align}
where, as introduced earlier, $\vx\in\R^n$ represent the primal decision variables, and $\vy\in\R^m$, $\vz\in\R^{p}$ corresponds to the Lagrange multipliers of equality and inequality constraints, respectively.
The necessary and sufficient conditions\footnote{In convex \gls{qp} problems, sufficiency follows from convexity of the objective and constraints, ensuring that the~\gls{kkt} conditions characterize optimality.} for optimality are given by the saddle point equation $\nabla\mathcal{L}(\vx,\vs,\vy,\vz) = \vzero$.
They are also known as first-order necessary conditions:
\begin{subequations}
\begin{align}
\matQ \vx + \vecC + \matA^\transpose \vy + \matG^\transpose \vz &= \vzero, \\
\matA \vx - \vecB &= \vzero, \\ 
\matG \vx + \vs - \vecH &= \vzero, \\
\vs \circ \vz = \vzero, (\vs, \vz) &\ge \vzero, \label{eq:complementarity_constraints}
\end{align}
\end{subequations}
where $\circ$ denotes an element-wise operation, which in this case represents the Hadamard product.

A central challenge in numerical optimization arises from the complementarity constraints described in~\Cref{eq:complementarity_constraints}.
This is because they require the solution to remain in the positive orthant.
Moreover, \gls{qp} problems can become degenerate, ill-posed, or large-scale, each of which introduces significant numerical difficulties.
Common sources of degeneration include:  
\begin{enumerate}[label=(\roman*)]
\item Rank-deficient constraint matrices: linear dependencies in $\matA$ or $\matG$, \label{enum:1_qp_challenges}
\item Non-uniqueness of dual solutions: multiple valid values for $\vy$ and $\vz$, \label{enum:2_qp_challenges}
\item Singular or semidefinite Hessians: lack of strict convexity in $\matQ$, \label{enum:3_qp_challenges}
\item Ill-conditioning from small constraint violations or near-zero multipliers: e.g., $\vz \approx \vzero$. \label{enum:4_qp_challenges}
\end{enumerate}
These issues make \gls{qp} problems difficult to solve reliably and efficiently with standard solvers. 
Complicating matters further, \gls{qp} problems may also be infeasible.

Given the impact of these numerical limitations on algorithms in robotics and AI, we design \textsc{Odyn} to tackle hard \gls{qp} problems from degenerate systems to warm-started sequences of related problems.
This makes it especially suitable for real-time applications in robotics and AI, where robustness and efficiency are critical.

Below, we delve into the key insights used in the design of the \textsc{Odyn} \gls{qp} solver.

\subsection{From interior to non-interior point}\label{subsec:ipm_to_nip}
In this section, we highlight the connections between \glspl{ipm} and \glspl{alm}, which motivate the development of our path-following \gls{nip} methods.
For clarity of exposition, we omit equality constraints; however, they can be easily integrated.

\glspl{ipm} relax the complementarity constraints by employing the log-barrier functions.
Inequality constraints are incorporated into the objective as barrier terms, yielding
\begin{equation}\label{eq:barrier_qp}
\begin{aligned}
\min\limits_{\vx, \vs} \quad
    & \tfrac{1}{2}\,\vx^\transpose \matQ\,\vx
      + \vecC^\transpose \vx
      - \mu \sum_{j=1}^{p} \log(\vs_j) \\
\text{subject to} \quad
    & \matG\,\vx + \vs = \vecH,
\end{aligned}
\end{equation}
where $\mu\in\R_{+}$ is the barrier parameter.
The associated~\gls{kkt} conditions can be written compactly as
\begin{equation}\label{eq:residual_ipm}
\vr(\U; \mu) =
\begin{bmatrix}
\matQ \vx + \vecC + \matG^\transpose \vz \\
\matG \vx + \vs - \vecH \\
\vs \circ \vz - \mu \,\bone
\end{bmatrix}
= \vzero,
\qquad
(\vs, \vz) \geq \vzero,
\end{equation}
where $\U = (\vx,\vs,\vz)$ denotes the collection of primal and dual variables.
To find the root of the above equation (i.e.,~\gls{kkt} point), we apply the Newton’s method to~\Cref{eq:residual_ipm}, which leads to the following linear system 
\begin{equation}\label{eq:newton_ipm_system}
\begin{bmatrix}
\matQ & \matG^\transpose &  \\
\matG &                  & \mathbf{I} \\
      & \matS            & \matZ
\end{bmatrix}
\begin{bmatrix}
\dx \\
\dz \\
\ds
\end{bmatrix}
=
-\,
\begin{bmatrix}
\matQ \vx + \vecC + \matG^\transpose \vz \\
\matG \vx + \vs - \vecH \\
\vs \circ \vz - \mu \,\bone
\end{bmatrix}
\end{equation}
with $\matS = \diag(\vs)$ and $\matZ = \diag(\vz)$ denoting the diagonal matrices of slack and dual variables, respectively, and $\bone$ as a vector with ones.

\Cref{eq:residual_ipm} satisfies a relaxed complementarity slackness condition, i.e., $\vs\circ{\vz} = \mu\bold{1}$.
When solving the parametrized \gls{qp} problem, each value of $\mu$ yields a primal--dual point along the central path.
In practice, we recover the solution of~\Cref{eq:slack_reformulated_qp} by solving a sequence of unconstrained problems with progressively decreasing barrier parameters, letting $\mu\rightarrow0$.

\begin{figure}[!t]
\centering
\includegraphics[width=2.5in]{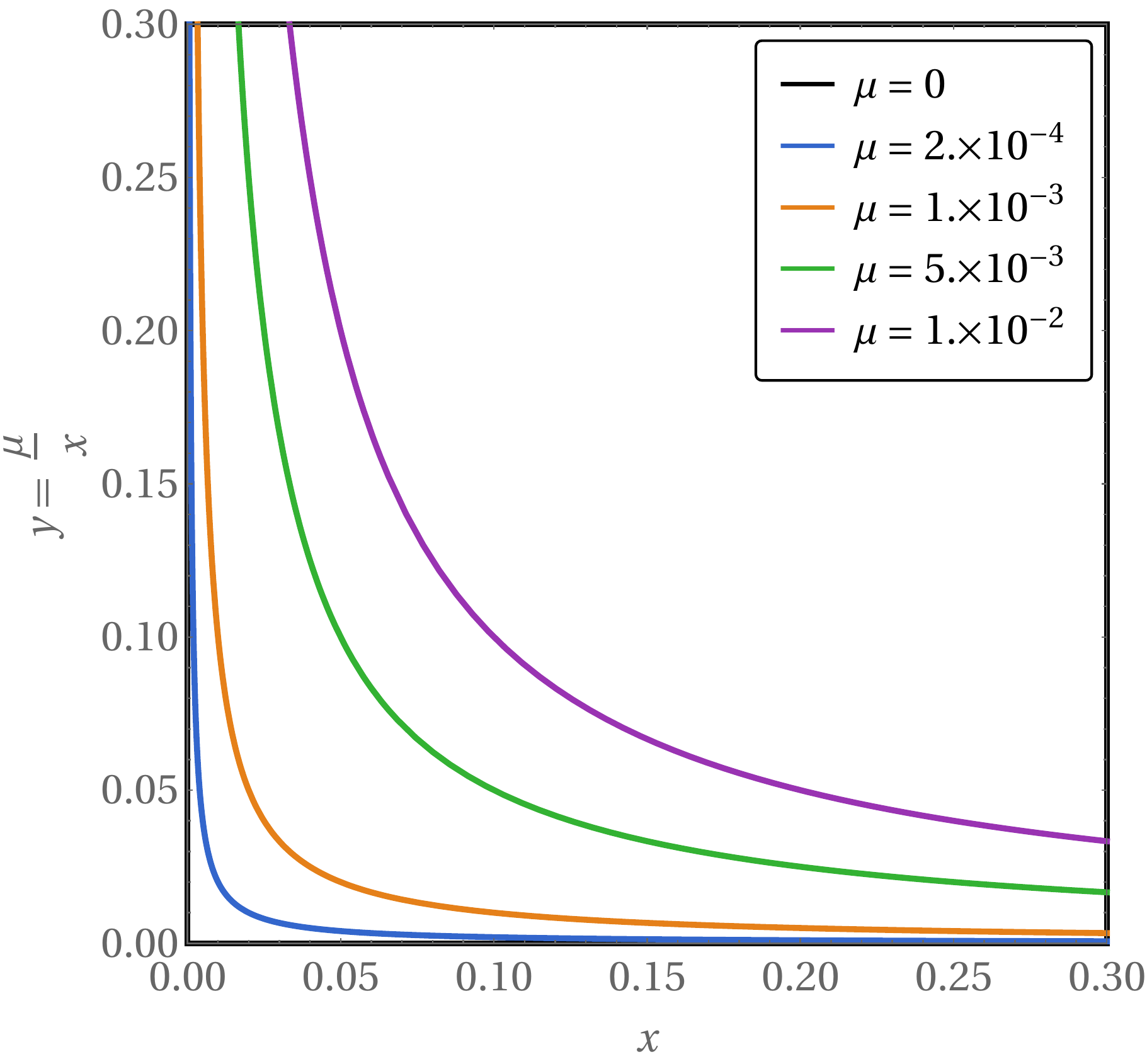}
\caption{Log-barrier function as $\mu$ approaches $0$.
The approximation becomes closer to the indicator function.}
\label{fig:log_barrier}
\end{figure}

Smoothing the complementarity conditions offers a more ``natural'' alternative to the projection mechanisms commonly used in \glspl{alm}, as it introduces a continuation procedure that gradually drives $\mu \to 0$ (see~\Cref{fig:log_barrier}).

However, logarithmic barriers force iterates to remain strictly within the interior of the feasible region. 
In addition to restricting the search direction, this constraint complicates warm-starting.
These limitations have motivated renewed interest in augmented-Lagrangian approaches for robotics and AI~\cite{bambade2022_proxqp}.
Nevertheless, we argue that their reliance on projection mechanisms can be restrictive.

To address the limitations inherent in both \glspl{ipm} and \glspl{alm}, we begin by introducing consensus variables $\vxi\in\R^p$:
\begin{equation}\label{eq:lifted_qp}
\begin{aligned}
\min\limits_{\vx, \vs, \vxi} \quad
    & \tfrac{1}{2}\,\vx^\transpose \matQ\,\vx + \vecC^\transpose \vx \\
\text{subject to} \quad
    & \matG\,\vx + \vs = \vecH, \\
    & \vs = \vxi, \\
    & \vxi \ge \vzero.
\end{aligned}
\end{equation}
We then construct an equality-constrained problem by applying log-barrier functions to the inequality constraints (as in \glspl{ipm}), i.e.,
\begin{equation}\label{eq:lifted_barrier_qp_problem}
\begin{aligned}
\min\limits_{\vx, \vs, \vxi} \quad
    & \tfrac{1}{2}\,\vx^\transpose \matQ\,\vx
      + \vecC^\transpose \vx
      - \mu \sum_{j=1}^{p} \log(\vxi_j) \\
\text{subject to} \quad
    & \matG\,\vx + \vs = \vecH, \\
    & \vs = \vxi.
\end{aligned}
\end{equation}
We finally relax the strict interior-point condition by introducing a quadratic penalty to the consensus constraint.
This yields to a \emph{penalty--barrier function} that blends ideas from \glspl{ipm} and \glspl{alm}:
\begin{equation}\label{eq:lifted_barrier_al_qp_problem}
\begin{aligned}
\min\limits_{\vx, \vs, \vxi} \quad
    & \tfrac{1}{2}\,\vx^\transpose \matQ\,\vx
      + \vecC^\transpose \vx
      - \mu \sum_{j=1}^{p} \log(\vxi_j)
      + \tfrac{1}{2}\,\|\vs - \vxi\|_2^2 \\
\text{subject to} \quad
    & \matG\,\vx + \vs = \vecH, \\
    & \vs = \vxi.
\end{aligned}
\end{equation}
The Lagrangian associated with this penalty--barrier formulation is given by
\begin{align}
\calP(\Ut; \mu)
&= \tfrac12 \vx^\transpose \matQ \vx
   + \vecC^\transpose \vx
   - \mu \sum_{j=1}^{p} \log(\vxi_j)
   + \tfrac12 \norm{\vs - \vxi}_2^2 \notag\\
&\quad - \vw^\transpose (\vs - \vxi)
   + \vz^\transpose (\matG \vx + \vs - \vecH),
\end{align}
where $\vw\in\R^p$ is the Lagrange multipliers associated with the consensus constraint and $\Ut = (\vx, \vs, \vxi, \vw, \vz)$ collects primal and dual decision variables.
Therefore, the first-order neccesary conditions are given by:
\begin{equation}\label{eq:novel_KKT_conditions}
\tilde{\vr}(\Ut; \mu)
=
\begin{bmatrix}
\nabla_{\vx} \calP \\[0.3em]
\nabla_{\vz} \calP \\[0.3em]
\nabla_{\vw} \calP \\[0.3em]
\nabla_{\vxi} \calP \\[0.3em]
\nabla_{\vs} \calP
\end{bmatrix}
=
\begin{bmatrix}
\matQ \vx + \vecC + \matG^\transpose \vz \\[0.3em]
\matG \vx + \vs - \vecH \\[0.3em]
\vs - \vxi \\[0.3em]
-\mu \,\vxi^{\circ -1} - (\vs - \vxi) + \vw \\[0.3em]
\vs - \vxi - \vw + \vz
\end{bmatrix}
= \bzero.
\end{equation}

From these conditions, we observe that we can find a closed-form solution for the consensus variable $\vxi$, as $\nabla_{\vxi}\calL = \bzero$ is a second-order polynomial, i.e., 
\begin{equation}\label{eq:xi_quadratic}
\vxi^{\circ 2}
\;-\;
(\vs - \vw) \circ \vxi
\;-\;
\mu\,\bone
= \vzero.
\end{equation}
To satisfy the condition $\vxi>\bzero$,  we must pick the positive solution:
\begin{equation}\label{eq:xi_quadratic_sol}
\vxi
=
\frac{
(\vs - \vw)
\;+\;
\sqrt{(\vs - \vw)^{\circ 2} + 4\mu\,\bone}
}{2}.
\end{equation}
We also observe that $\vw = \vz$ from equations $\nabla_{\vw}\calL = \vzero$ and $\nabla_{\vs}\calL = \vzero$.
This leads to: 
\begin{equation}
\vxi
=
\frac{
(\vs - \vz)
\;+\;
\sqrt{(\vs - \vz)^{\circ 2} + 4\mu\,\bone}
}{2}.
\end{equation}
Finally, by injecting the above result into $\nabla_{\vw}\calL = \vzero$ we obtain the equivalent~\gls{kkt} conditions:
\begin{equation}\label{eq:min_nip_KKT_conditions}
\vr(\U; \mu)
=
\begin{bmatrix}
\matQ \vx + \vecC + \matG^\transpose \vz \\[0.4em]
\matG \vx + \vs - \vecH \\[0.4em]
\vs + \vz - \sqrt{(\vs - \vz)^{\circ 2} + 4\mu\,\bone}
\end{bmatrix}
= \vzero.
\end{equation}
\Cref{eq:min_nip_KKT_conditions} can be rewritten as follows:
\begin{equation}\label{eq:nip_KKT_conditions}
\vr(\U; \mu)
=
\begin{bmatrix}
\matQ \vx + \vecC + \matG^\transpose \vz \\[0.3em]
\matG \vx + \vs - \vecH \\[0.3em]
\vncpFun{}(\vs, \vz; \mu)
\end{bmatrix}
=
\vzero,
\end{equation}
where $\vncpFun{} : \R^{p} \times \R^{p} \to \R^{p}$ is a general~\gls{ncp} function, representing $\vncpFun{\mathrm{MIN}}$ or $\vncpFun{\mathrm{FB}}$.
This is because we could recognize the smoothed minimum~\gls{ncp} function $\vncpFun{\mathrm{MIN}} : \R^{p} \times \R^{p} \to \R^{p}$, i.e.,
\begin{equation}\label{eq:weight_ncp_function}
\vncpFun{\mathrm{MIN}}(\vs, \vz; \mu)
=
\vs + \vz - \sqrt{(\vs - \vz)^{\circ 2} + 4\mu\,\bone}.
\end{equation}
Moreover, smoothed~\gls{ncp} functions satisfy the following fundamental property~\cite{kanzow1996}:
\begin{equation}
\label{eq:ncp_property}
\vncpFun{}(\vs, \vz; \mu) = \bzero
\;\Longleftrightarrow\;
\vs \ge \vzero,\quad
\vz \ge \vzero,\quad
\vs \circ \vz = \mu\,\bone.
\end{equation}
where, by injecting the centrality constraint $\vs\circ{\vz} = \mu\bone$, we transform a smoothed minimum~\gls{ncp} function as a regularized Fischer--Burmeister~\gls{ncp}, i.e.,
\begin{align}
\vncpFun{\mathrm{MIN}}(\vs, \vz; \mu)
  &= \vs + \vz
     - \sqrt{\,\vs^{\circ 2} - 2\,\vs \circ \vz + \vz^{\circ 2} + 4\mu\,\bone\,} \notag \\[0.3em]
  &= \vs + \vz
     - \sqrt{\,\vs^{\circ 2} + \vz^{\circ 2} + 2\mu\,\bone\,} \notag \\[0.3em]
  &= \vncpFun{\mathrm{FB}}(\vs, \vz; \mu). \label{eq:phi_min}
\end{align}

Finally, the nonlinear system of equations in~\Cref{eq:nip_KKT_conditions} is solved by applying Newton's method:
\begin{equation}
\begin{bmatrix}
\matQ & \matG^\transpose &  \\
\matG &                  & \Id \\
      & \ncpJac{\vz}       & \ncpJac{\vs}
\end{bmatrix}
\begin{bmatrix}
\dx \\
\dz \\
\ds
\end{bmatrix}
=
-
\begin{bmatrix}
\matQ \vx + \vecC + \matG^\transpose \vz \\
\matG \vx + \vs - \vecH \\
\vncpFun{}(\vs,\vz;\mu)
\end{bmatrix},
\end{equation}
where $\ncpJac{\vz}\in\R^{p\times p}$ and $\ncpJac{\vs}\in\R^{p\times p}$ are diagonal matrices denoting the~\gls{ncp} Jacobians.

To promote convergence, the iterates of \gls{nip} methods are typically constrained to a neighborhood of the central path.
This is defined (see~\cite{huang2003predictor}) as follows:
\begin{equation}\label{eq:nip_neighborhood}
\NipNbr{\beta}
=
\left\{
\U \in \R^{\,n + 2p}
:\;
\norm{\vr(\U;\mu)} \;\le\; \beta\,\mu,
\;\mu > 0
\right\},
\end{equation}
where $\beta$ determines the radius of the neighborhood, ensuring that the iterates remain sufficiently close to the central path as $\mu\rightarrow0$.
Compared to \glspl{ipm}, this approach does not impose restrictions on $\vs$ and $\vz$, since the domain of $\vncpFun{}(\vs,\vz;\mu)$ includes the positive orthant of these variables.
Next, we delve into another insight on which \textsc{Odyn} is based.

\subsection{From degeneracy to regularity}
We begin by understanding what degeneracy is.
To do so, we focus on equality-constrained \gls{qp} problems:
\begin{equation}\label{eq:equality_qp}
\begin{aligned}
\min\limits_{\vx \in \mathbb{R}^n} \quad
    & \tfrac{1}{2}\,\vx^\transpose \matQ\,\vx + \vecC^\transpose \vx \\
\text{subject to} \quad
    & \matA\,\vx = \vecB.
\end{aligned}
\end{equation}
where its~\gls{kkt} system is defined as:
\begin{equation}
\begin{bmatrix}
\matQ & \matA^\transpose \\
\matA & 
\end{bmatrix}
\begin{bmatrix}
\dx \\
\dy
\end{bmatrix}
=
-
\begin{bmatrix}
\matQ \vx + \vecC + \matA^\transpose \vy \\
\matA \vx - \vecB
\end{bmatrix}.
\end{equation}

To be able to factorize this~\gls{kkt} system, we require (i) strong convexity ($\matQ\succ \bzero$) and (ii) linear independence in the constraint matrix $\matA$ ($\matA$ is full rank).
Handling this becomes more challenging if the~\gls{kkt} matrix is ill-conditioned.
For instance, apart of the extra computational demand, pseudo-inverses are not an option because:
\begin{enumerate}[label=(\roman*)]
    \item Numerical instability occurs when $\matQ$ or $\matA$ are ill-conditioned or rank deficient. 
    \item Their \emph{one-shot projection} (i.e., poor globalization) does not provides a convergence mechanism or guarantee.
\end{enumerate} 

Proximal point algorithms introduce regularity by relying on quadratic penalties~\cite{rockafellar1976_ppa}.
They can be used to handle \gls{qp} problems that are not strict convex or with ill-conditioning by casting the problem as follows
\begin{equation}\label{eq:equality_proximal_point}
\begin{aligned}
\max_{\vx}\min_{\vy} &\;\;
\calL(\vx, \vy)
+
\tfrac{\rhod}{2}\norm{\vx - \xE}_2^{2}
-
\tfrac{\rhoe}{2}\norm{\vy - \yE}_2^{2},
\end{aligned}
\end{equation}
where $\calL(\vx, \vy) = \tfrac12\, \vx^\transpose \matQ \vx + \vecC^\transpose \vx + \vy^\transpose(\matA \vx - \vecB)$, and $\xE\in\R^n$ and $\yE\in\R^m$ are the primal and dual estimates, respectively, and $\rhod,\rhoe\in\R_+$ are the proximal penalties.
\Cref{eq:equality_proximal_point} is equivalent to shifting the constraints by ($\xE$, $\yE$).
This is because~\Cref{eq:equality_proximal_point} can be rewritten as:
\begin{align}\label{eq:reformulated_equality_proximal_point}
\min_{\vx, \vy}\;\;
& 
\tfrac12\vx^\transpose \matQ \vx
+ \vecC^\transpose \vx
+ \tfrac{\rhod}{2}\norm{\vx - \xE}_2^{2}
+ (\matA \vx - \vecB)^\transpose \yE \notag\\[0.3em]
&\quad
+ \frac{1}{2\rhoe}\norm{\matA \vx - \vecB}_2^{2}
-
\frac{1}{2\rhoe}
\norm{\matA \vx - \vecB - \rhoe(\vy - \yE)}_2^{2}.
\end{align}

From the problem in~\Cref{eq:reformulated_equality_proximal_point}, we have the following \emph{perturbed~\gls{kkt} conditions}:
\begin{equation}
\vr(\hat{\U};\, \hat{\boldsymbol{\rho}},\, \hat{\U}_E)
=
\begin{bmatrix}
\matQ \vx + \vecC + \matA^\transpose \vy + \rhod(\vx - \xE) \\[0.4em]
\matA \vx - \vecB - \rhoe(\vy - \yE)
\end{bmatrix},
\end{equation}
where $\hat{\U} = (\vx, \vy)$, $\hat{\U}_E = (\xE, \yE)$, $\hat{\boldsymbol{\rho}} = (\rho_d, \rho_e)$ denote the stacked primal--dual variables, their estimates, and the regularization parameters, respectively.

The last couple of terms of~\Cref{eq:reformulated_equality_proximal_point} can be seen as the \emph{Forsgren--Gill primal--dual penalty function} associated with the shifted constraints: 
\begin{equation}
\matA \vx - \vecB - \rhoe\,\yE = \bzero.
\end{equation}
This is also know as proximal point Lagrangian function. Additionally, the term $\tfrac{\rhod}{2}\norm{\vx - \xE}_2^{2}$ in~\Cref{eq:reformulated_equality_proximal_point} can be recognized as the proximal point penalty function.
Therefore, combining both proximal point functions together with the term $\yE^\transpose(\matA \vx - \vecB)$ lead to a \emph{shifted primal--dual penalty function}, or a primal--dual proximal point penalty and Lagrangian function.

The parameters $\rho_d$, $\rho_e$, $\xE$ and $\yE$ provide a central path towards the solution.
This defines a convergence mechanism in contrast to pseudo-inverses.
Finally, we now need to factorize the following perturbed~\gls{kkt} system:
\begin{equation}
\begin{bmatrix}
\Qtil & \matA^\transpose \\[0.3em]
\matA & -\rhoe\mtr{I}
\end{bmatrix}
\begin{bmatrix}
\dx \\[0.3em]
\dy
\end{bmatrix}
=
-
\begin{bmatrix}
\matQ \vx + \vecC + \matA^\transpose \vy + \rhod(\vx - \xE) \\[0.4em]
\matA \vx - \vecB - \rhoe(\vy - \yE)
\end{bmatrix}
\end{equation}
with $\Qtil = \matQ + \rhod\mtr{I}$ denoting the regularized Hessian matrix, which ensures that the sufficient optimality conditions are satisfied even for linear programs and quasi-convex problems.

We are now ready to describe how degeneracy in complementarity conditions is handled via an all-shifted mechanism, and to introduce the \textsc{Odyn} \gls{qp} algorithm.

\section{All-shifted non-interior point QP (Odyn)}
This section presents the mathematical formulation and algorithmic design of \textsc{Odyn}. We start by describing our novel all-shifted~\gls{ncp} functions.

\subsection{All-shifted~\gls{ncp} functions}
To provide regularity to the nonlinear complementarity conditions and effectively handle degeneracy, we formulate~\Cref{eq:lifted_barrier_al_qp_problem} as:
\begin{equation}\label{eq:all_shifted_ncp_lagrangian}
\begin{aligned}
\max_{\vx, \vs, \vxi}\min_{\vw, \vz}\;\;
& \calP(\Ut; \mu)
+ \frac{\rhon}{4}\norm{\vs - \sE}_2^{2} \\[0.3em]
&- \frac{\rhon}{4}\norm{\vw - \wE}_2^{2}
- \frac{\rhon}{2}(\vs - \sE)^\transpose(\vw - \wE),
\end{aligned}
\end{equation}
where $\sE\in\R^p$ and $\wE\in\R^p$ are the estimates of the slack variables and consensus multipliers, respectively.
Note that we recover the optimality conditions of~\Cref{eq:lifted_barrier_al_qp_problem} when $\sE = \vs^{*}$ and $\wE = \vw^{*}$ and, for the sake of simplicity, we restrict our attention to inequality constraints.

\Cref{eq:all_shifted_ncp_lagrangian} yields to the following \emph{perturbed~\gls{ncp} Lagrangian}:
\begin{equation}
\begin{aligned}
\tilde{\calP}(\tilde{\mathbf U};\mu,\rho_n,\mathbf s_E,\mathbf w_E)
&= \calP(\tilde{\mathbf U};\mu) + \frac{\rho_n}{4}\| \mathbf s - \mathbf s_E\|_2^2 \\
&\hspace{-5em} {}- \frac{\rho_n}{4}\|\mathbf w - \mathbf w_E\|_2^2
  - \frac{\rho_n}{2}(\mathbf s - \mathbf s_E)^{\transpose}(\mathbf w - \mathbf w_E)
\end{aligned}
\end{equation}
in which its first-order necessary conditions are obtained by taking the following derivatives:
\begin{equation}\label{eq:all_shifted_KKT_conditions}
\begin{aligned}
\nabla_{\vx}\,\tilde{\calP}
&= \matQ \vx + \vecC + \matG^\transpose \vz, \\[0.3em]
\nabla_{\vy}\,\tilde{\calP}
&= \matG \vx + \vs - \vecH, \\[0.3em]
\nabla_{\vw}\,\tilde{\calP}
&= \vs - \vxi
   + \frac{\rhon}{2}(\vs - \sE)
   + \frac{\rhon}{2}(\vw - \wE), \\[0.3em]
\nabla_{\vxi}\,\tilde{\calP}
&= -\,\mu\,\vxi^{-1}
   - (\vs - \vxi)
   + \vw, \\[0.3em]
\nabla_{\vs}\,\tilde{\calP}
&= \vs - \vxi
   + \frac{\rhon}{2}(\vs - \sE)
   - \frac{\rhon}{2}(\vw - \wE)
   - \vw + \vz.
\end{aligned}
\end{equation}
Similarly to~\Cref{subsec:ipm_to_nip}, we find a closed-form solution for the consensus variable $\vxi$.
Additionally, we observe that~\Cref{eq:all_shifted_KKT_conditions} is the perturbed version of~\Cref{eq:novel_KKT_conditions}.

From $\nabla_{\vw}\tilde{\calP} = \vzero$ and $\nabla_{\vs}\tilde{\calP} = \vzero$, we obtain a perturbed dual consensus constraint $\vz - \vw - \rho_n(\vw - \wE)$.
Therefore, the first-order necessary conditions in~\Cref{eq:all_shifted_KKT_conditions} can be equivalently rewritten as:
\begin{equation}\label{eq:all_shifted_KKT_residual}
\vr(\U;\mu) =
\begin{bmatrix}
\matQ \vx + \vecC + \matG^\transpose \vz \\[0.4em]
\matG \vx + \vs - \vecH \\[0.4em]
\vz - \vw - \rhon(\vw - \wE) \\[0.4em]
\vncpFun{}(\vs, \vw;\mu)
+ \rhon(\vs - \sE)
+ \rhon(\vw - \wE)
\end{bmatrix},
\end{equation}
in which, by substituting the solution $\vxi$ in $\nabla_{\vw}\calL = \vzero$, the all-shifted~\gls{ncp} functions is obtained as follows
\begin{equation}\label{eq:all-shifted-ncp}
\vncpFun{}(\vs,\vw;\mu)
+ \rhon(\vs - \sE)
+ \rhon(\vw - \wE).
\end{equation}

Including this regularity helps dealing with degenerated complementarity conditions, i.e., for some $j$ we may have $\vs_j = \vz_j = 0$.
For instance, if we do not include this regularization, the~\gls{ncp} becomes flat nears convergence (i.e, $\mu = 0$). Additionally, similar to \Cref{eq:ncp_property} the all-shifted NCP functions satisfy the following property:
\begin{equation}
\vncpFun{}(\vs, \vw; \mu) = \bzero
\;\Longleftrightarrow\;
\begin{cases}
\vs \ge \vzero, \quad \vw \ge \vzero, \\[4pt]
\vs \circ \vw
= \mu\,\bone
+ \vRhoBar_n \circ
  \bigl(
    \sE - \vs
    + \wE - \vw
  \bigr),
\end{cases}
\end{equation}

\noindent
where $\vRhoBar_n = \frac{\rho_n}{2}\vs$.
This property is demonstrated in~\Cref{app:appendix}.

We describe a complete algorithm for solving degenerated \gls{qp} problems in the next section.

\subsection{All-shifted non-interior quadratic programming}
We combine the all-shifted \gls{ncp} functions with proximal penalties and Lagrangian terms to develop our path-following \gls{nip} method.
In particular, \textsc{Odyn} solves the following perturbed saddle-point problem:
\begin{equation}\label{eq:odyn_lagrangian}
\begin{aligned}
\max_{\vx, \vs, \vxi}\;\min_{\vw, \vy, \vz}\quad
& \tilde{\calP}(\Ut; \mu, \rhon, \sE, \wE)
+ \frac{\rhod}{2}\norm{\vx - \xE}_2^{2} \\[0.3em]
&\quad
- \frac{\rhoe}{2}\norm{\vy - \yE}_2^{2}
- \frac{\rhoi}{2}\norm{\vz - \zE}_2^{2},
\end{aligned}
\end{equation}
where $\rho_i\in\R_+$ and $\zE\in\R^p$ are the inequality regularization parameter and estimates, respectively, and including inequality constraints in $\tilde{\calP}(\Ut; \mu, \rhon, \sE, \wE)$ with $\Ut = (\vx, \vs, \vxi, \vw, \vy, \vz)$. 
Its~\gls{kkt} conditions are given by:
\begin{equation}\label{eq:odyn_residual}
\begin{bmatrix}
\tilde{\vr}_d \\[0.5em]
\tilde{\vr}_e \\[0.5em]
\tilde{\vr}_i \\[0.5em]
\tilde{\vr}_c \\[0.5em]
\tilde{\vr}_g
\end{bmatrix}
=
\begin{bmatrix}
\matQ \vx + \vecC + \matA^\transpose \vy + \matG^\transpose \vz + \rhod(\vx - \xE) \\[0.5em]
\matA \vx - \vecB - \rhoe(\vy - \yE) \\[0.5em]
\matG \vx + \vs - \vecH - \rhoi(\vz - \zE) \\[0.5em]
\vz - \vw - \rhon(\vw - \wE) \\[0.5em]
\vncpFun{}(\vs, \vw;\mu) \;+\; \rhon(\vs - \sE) + \rhon(\vw - \wE)
\end{bmatrix},
\end{equation}
where $\tilde{\vr}_d\in\R^n$ is the perturbed dual residual, $\tilde{\vr}_e\in\R^m$, $\tilde{\vr}_i\in\R^p$ and $\tilde{\vr}_c\in\R^p$ are the perturbed residual vectors associated to equality, inequality and consensus constraints, respectively.

In practice, we observe that the proximal term associated with the consensus constraint $\tilde{\vr}_c$ has a negligible effect. 
Therefore, we further condense the \gls{kkt} conditions by substituting $\vw = \vz$ into \Cref{eq:odyn_residual}, yielding:
\begin{equation}\label{eq:odyn_residual}
\tilde{\vr}(\U;\boldsymbol{\Pi}) = 
\begin{bmatrix}
\tilde{\vr}_d\\
\tilde{\vr}_e\\
\tilde{\vr}_i\\
\tilde{\vr}_g
\end{bmatrix}
=
\begin{bmatrix}
\matQ \vx + \vecC + \matA^\transpose \vy + \matG^\transpose \vz + \rhod(\vx - \xE)\\
\matA \vx - \vecB - \rhoe(\vy - \yE)\\
\matG \vx + \vs - \vecH - \rhoi(\vz - \zE)\\
\vncpFun{}(\vs,\vz;\mu) + \rhon(\vs - \sE) + \rhon(\vz - \zE)
\end{bmatrix}
\end{equation}

\subsection{Search direction}
To compute the~\gls{kkt} point in~\Cref{eq:odyn_residual}, we apply the Newton's method.
As the Newton's method iterates over a linear approximation of the~\gls{ncp} functions, this produces a search direction at each iteration obtained by factorizing the following saddle-point system:
\begin{equation}\label{eq:full_kkt_system}
\begin{bmatrix}
\Qtil      & \matA^\transpose & \matG^\transpose &  \\
\matA      & -\rhoe \Id &           &  \\
\matG      &            & -\rhoi \Id & \Id \\
           &            & \pertncpJac{\vz} & \pertncpJac{\vs}
\end{bmatrix}
\begin{bmatrix}
\dx\\
\dy\\
\dz\\
\ds
\end{bmatrix}
=
-
\begin{bmatrix}
\tilde{\vr}_d\\
\tilde{\vr}_e\\
\tilde{\vr}_i\\
\tilde{\vr}_g
\end{bmatrix},
\end{equation}
where $\pertncpJac{\vs} = \ncpJac{\vs} + \rho_{n}\Id$ and $\pertncpJac{\vz} = \ncpJac{\vz} + \rho_{n}\Id$ are the regularized~\gls{ncp} Jacobians with respect to the slack variables and inequality multipliers, respectively.  

To increase computational efficiency, we could condense~\Cref{eq:full_kkt_system} as follows:
\begin{equation}\label{eq:reduced_kkt_system}
\begin{bmatrix}
\Qtil & \matJ^\transpose \\[0.3em]
\matJ  & -\,\matD
\end{bmatrix}
\begin{bmatrix}
\dx \\[0.3em]
\dl
\end{bmatrix}
=
-
\begin{bmatrix}
\tilde{\vr}_d \\[0.3em]
\tilde{\vr}_p
\end{bmatrix},
\end{equation}
where $\dl = \begin{bmatrix} \dy^\transpose & \dz^\transpose \end{bmatrix}^{\!\transpose}\in\R^{m+p}$ is the stacked search direction for equality and inequality multipliers, $\tilde{\vr}_{p}
= \begin{bmatrix} \tilde{\vr}_{e}^{\!\transpose} & \bar{\vr}_{i}^{\!\transpose} \end{bmatrix}^{\!\transpose}\in\R^{m+p}$ is the right-hand-side term associated to the constraint residuals (i.e., the perturbed residuals), $\Bar{\vr}_i = \tilde{\vr}_i - \vq$ is the condensed right-side term used to compute $\dz$, $\matP = \pertncpJac{\vs}^{-1}\,\pertncpJac{\vz}$ is a diagonal matrix, $\vq = \pertncpJac{\vs}^{-1}\tilde{\vr}_g$, $\matJ^\transpose = \begin{bmatrix} \matA^\transpose & \matG^\transpose \end{bmatrix}^\transpose$ stacks the Jacobian matrices of the constraints, and the diagonal matrix $\matD$ is defined as: 
\begin{equation}
\matD =
\begin{bmatrix}
\rhoe\,\Id &  \\[0.3em]
 & \tilde{\matP}
\end{bmatrix},
\end{equation}
with $\tilde{\matP} = \matP + \rhoi\Id$.
The system in~\Cref{eq:reduced_kkt_system} can be factorized using a condensed Hessian matrix $\matH = \Qtil + \matJ^\transpose\matD^{-1}\matJ\succ\bzero$.
Since $\matH$ is positive definite, we employ a Cholesky factorization, exploiting the trivial inversion of the diagonal matrix~$\matD$. 
This yields the following search directions:
\begin{subequations}\label{eq:odyn_search_direction}
\begin{align}
\dx
  &= -\,\matH^{-1}\!\left(\vtresd + \matJ^{\!\transpose}\matD^{-1}\vtresp\right), \\[0.5em]
\dl
  &= \matD^{-1}\!\left(\matJ\,\dx + \vtresp\right), \\[0.5em]
\ds
  &= -\left(\matP\,\Delta\vz + \vq\right).
\end{align}
\end{subequations}
We refer to this approach as the \emph{condensed-\gls{kkt}} formulation. 
In contrast, the \emph{full-\gls{kkt}} system in~\Cref{eq:full_kkt_system} can be factorized using an LDL$^\top$ decomposition or a quasi-definite LDT scheme~\cite{golub-matcompbook}. 
We now describe the step-acceptance criteria.

\subsection{Step acceptance}
A candidate solution is obtained from $\Delta\U = (\Delta\vx,  \Delta\vs, \Delta\vy, \Delta\vz)$ via a line search procedure, i.e, $\U^+ = \U + \alpha\Delta{\U}$, where $\alpha\in(0, 1]$ is the step length parameter.
To assess a candidate solution, we employ a natural merit function defined as follows:
\begin{equation}
\NatMerit(\U) = \frac{1}{2}(\tilde{\vr}_d^\transpose\tilde{\vr}_d + \tilde{\vr}_p^\transpose\tilde{\vr}_p + \tilde{\vr}_g^\transpose\tilde{\vr}_g),
\end{equation}
where $\tilde{\vr}_p = \begin{bmatrix}\tilde{\vr}_e^\transpose & \tilde{\vr}_i^\transpose\end{bmatrix}^\transpose$ is the concatenation of the perturbed equality and inequality constraint residuals.

Although \gls{ncp} functions temporarily allow iterates to leave the positive orthant, the feasible neighborhood they define is typically narrow, limiting the step lengths accepted during line search.
To mitigate this limitation, we develop a \textit{flexible backtracking line search} based on a relaxed Armijo condition. Specifically, for a candidate step length $\alpha$, we require:
\begin{equation}\label{proposed_armijo_equations}
\NatMerit\!\left(\U + \alphals\,\dU\right)
\;\le\;
\gammaLS\,\NatMerit(\U)
\;+\;
\alphals\,\etaLS\;
D\NatMerit(\U)^{\!\transpose}\dU,
\end{equation}

\noindent
where $\gamma>0$ is a scaling parameter that controls the degree of non-monotonicity and $\eta>0$ is a small positive constant.

This condition draws inspiration from the non-monotone line search in~\cite{zhang2004nonmonotone}, but differs in that we use a scaled version of the current merit function as the reference value rather than a moving average. 
Additionally, the natural merit function $\NatMerit$ is evaluated using the perturbed \gls{kkt} residuals.

\subsection{Centering algorithm}
The barrier or smoothing parameter for the~\gls{ncp} functions is computed as proposed in~\cite{kanzow1996}, i.e.,
\begin{equation}
\mu = \frac{\|\min(\vs, \vz)\|_2^2}{p},
\end{equation}
where $p$ is the number of inequality constraints.
This formulation ensures that $\mu$ reflects the degree of complementarity violation.
The smoothing parameter is updated using:
\begin{equation}
\mu^{+} = \max(\mu_{\min}, \sigma\mu),
\end{equation}
where $\sigma\in(0, 1)$ is a reduction factor, often referred to as the \textit{centering parameter} in the~\gls{ipm} literature \cite{nocedal-optbook}.

\begin{algorithm}[b]
    \tcc{compute predictor for the complementarity measure}
    $\mu \gets \|\min(\vs,\, \vz)\|^2 / p$\\
    $\mu \gets \max(\mu_{\min},\, \sigma \mu)$
    \tcc{update centering parameter}
    \If{$\|\vr(\U^+;\boldsymbol{\Pi})\| \leq \theta_u \|\vr(\U;\boldsymbol{\Pi})\|$}{
        $\mu \gets \delta_\mu^+\, \mu$\\
        $\sigma \gets \max(\sigma_{\min},\, \sigma - \delta^{-}\sigma)$
    }
    \Else{
        $\sigma \gets \min(\sigma_{\max},\, \sigma + \delta^{+}(1 - \sigma))$
    }
    \tcc{safeguard against growth}
    $\mu \gets \min(\mu,\, \mu_{\text{old}})$

    \caption{Centering algorithm}
    \label{alg:centering_algorithm}
\end{algorithm}

The centering parameter $\sigma$ is adjusted dynamically according to a \emph{trust-region–inspired rule}, as detailed in~\Cref{alg:centering_algorithm}. 
Our strategy resembles the heuristic used in \textsc{PiQP}~\cite{schwan2023_piqp} for updating proximal regularization parameters. 
However, our update is less aggressive: we apply fixed decrease and increase factors based on the observed progress. 
Specifically, if the \gls{kkt} residual $\vr^{+}$ is sufficiently reduced relative to the previous residual $\vr$, we decrease the centering parameter as
\begin{equation}
\sigma^{+} = \max\!\left(\sigma_{\min},\ \sigma - \delta^{-}\sigma\right),
\end{equation}
or increase the centering parameters to promote larger steps as
\begin{equation}
\sigma^+ = \min(\sigma_{max},\ \sigma + \delta^{+}(1 - \sigma)),
\end{equation}
where $\sigma_{\min}$ and $\sigma_{\max}$ are the minimum and maximum allowed value for the centering parameter, respectively, with $\delta^-$ and $\delta^+$ as the decrease and increase factors.

In~\Cref{alg:centering_algorithm}, the barrier parameter $\mu$ is enforced to decrease monotonically. 
If an increase in the complementarity residual is detected, $\mu$ is clamped to its previous value $\mu_{\textrm{old}}$ to prevent growth in the barrier parameter.
Note that $\boldsymbol{\Pi} = (\mu, \boldsymbol{\rho}, \mathbf{U}_E)$ denotes the collection of \textsc{Odyn} parameters

\subsection{Proximal primal--dual estimates}
Proximal methods typically use either the previous or the current primal--dual iterate as the reference point for the proximal terms~\cite{bambade2022_proxqp, schwan2023_piqp}. 
In contrast, we introduce an interpolated estimate between these two choices. 
Specifically, the updated primal--dual references are computed as:
\begin{align}\nonumber\label{eq_proximal_primal_dual_estimates}
\xE &= \vx^+ + \theta_{d}(\vx - \vx^{+}), &
\yE &= \vy^+ + \theta_{e}(\vy - \vy^{+}), \\
\zE &= \vz^+ + \theta_{i}(\vz - \vz^{+}), &
\sE &= \vs^+ + \theta_{i}(\vs - \vs^{+}),
\end{align}
where $\theta_d, \theta_e, \theta_i\in[0,1]$ are interpolation parameters for dual, equality, inequality, and~\gls{ncp} feasibility, respectively.
These interpolation parameters are updated dynamically using a trust-region-inspired strategy.   
When a sufficient reduction in relative feasibility is observed (i.e., measured against a threshold factor $\theta_l$), the corresponding interpolation parameter is decreased exponentially.
The general update rule for each $\theta_l$ is given by:  
\begin{equation}
\theta_l = \mathcal{T}(\theta_l - \theta^{-}\theta_l; \theta_{\min}),
\end{equation}
where $\theta^{-} > 0$ is the decay factor and $\mathcal{T}(x;t)$ is a hard threshold function that sets its input $x$ to zero if it falls below a minimum threshold $t$.
Conversely, if the relative feasibility degrades beyond an upper threshold factor $\theta_u$, the interpolation parameter is increased:
\begin{equation}
\theta_l = \min(1, \theta_l + \theta^{+}(1 - \theta_l)),
\end{equation}
where $\theta^{+} > 0$ is the growth factor. 
The update logic is summarized in~\Cref{alg_compute_step_length_estimates}.

\begin{algorithm}[b]
    \tcc{dual step length and regularization update}
    \If{$\|\vr_d^+\|_\infty \leq \theta_l \|\vr_d\|_\infty$}{
        $\theta_d \gets \mathcal{T}(\max(0,\, \theta_d - \theta^{-}\theta_d);\theta_{\min})$\\
        $\rho_d \gets \min(\rho_d \cdot \delta^{-1},\, \rho_{d,\min})$
    }
    \ElseIf{$\|\vr_d^+\|_{\infty} \geq \theta_u \|\vr_d\|_{\infty}$}{
        $\theta_d \gets \min(1,\, \theta_d + \theta^{+}(1 - \theta_d))$
    }
    \tcc{primal step length and regularization update}
    \If{$\|\vr_p^+\|_\infty \leq \theta_l \|\vr_p\|_\infty$}{
        $\theta_p \gets \mathcal{T}(\max(0,\, \theta_p - \theta^{-}\theta_p);\theta_{\min})$\\
        $\rho_p \gets \min(\rho_p \cdot \delta^{-1},\, \rho_{p,\min})$
    }
    \ElseIf{$\|\vr_p^+\|_\infty \geq \theta_u \|\vr_p\|_\infty$}{
        $\theta_p \gets \min(1,\, \theta_p + \theta^{+}(1 - \theta_p))$
    }
    \tcc{\gls{ncp} regularization update}
    \If{$\|\vr_g^+\|_\infty \leq \theta_l \|\vr_g\|_\infty$}{
        $\rho_n \gets \min(\rho_n \cdot \delta^{-1},\, \rho_{n,\min})$
    }
    \caption{Compute step length estimates}
    \label{alg_compute_step_length_estimates}
\end{algorithm}

\subsection{Neighborhood termination criteria}
The neighborhood condition typically used in \gls{nip} algorithms (see~\Cref{eq:nip_neighborhood}) may require several Newton iterations to be satisfied before the smoothing parameter can be reduced. 
With hard \gls{qp} problems, however, this condition can become difficult to meet.
This is because it does not account for the scale of the problem and effectively imposes an absolute tolerance on a theoretical subproblem.

To address this, we solve the perturbed primal--dual \gls{kkt} system using fixed values of the smoothing parameter $\mu$ and the regularization terms introduced in~\Cref{eq:reduced_kkt_system}. 
Unlike primal--dual~\glspl{alm} such as \textsc{ProxQP}~\cite{bambade2022_proxqp}, which rely on an absolute tolerance to terminate the parameterized subproblem, we instead employ a relaxed \emph{relative} neighborhood condition defined as:
\begin{equation}\label{neighborhood_termination_critera}
\begin{aligned}
\calN(\tau) = \Bigl\{ \U:\| \tilde{\vr}(\U^{+\alpha}; \boldsymbol{\Pi}) 
&\|_\infty \leq \theta \| \tilde{\vr}(\U;\boldsymbol{\Pi}) \|_\infty + \beta\mu \Bigr\},
\end{aligned}
\end{equation}
where $\boldsymbol{\tau} = (\beta, \theta)$, $\boldsymbol{\Pi} = (\mu, \boldsymbol{\rho}, \mathbf{U}_E)$ denotes the collection of \textsc{Odyn} parameters, $\beta > 0$ is a constant, and $\theta \in (0,1)$ specifies the required relative decrease in the perturbed residual norm. 
When the condition in~\Cref{neighborhood_termination_critera} is satisfied, we update the smoothing and regularization parameters as well as the proximal primal--dual estimates.

\subsection{Stopping criteria}
Inspired by \textsc{PiQP}~\cite{schwan2023_piqp}, we employ a termination criterion that combines relative and absolute tolerances. 
The \textsc{Odyn} \gls{qp} solver terminates when the primal and dual residuals fall below a threshold defined as the sum of an absolute tolerance $\epsilon_a > 0$ and a relative tolerance $\epsilon_r \ge 0$ scaled by the problem data. 
Specifically, we require:
\begin{equation}\label{eq:stopping_criteria}
\begin{aligned}
\|\vr_d\|_\infty 
  &\leq \epsilon_a 
    + \epsilon_r \max\!\big(
        \|\matQ \vx\|_\infty, 
        \|\vecC\|_\infty,
        \|\matA^\top \vy\|_\infty, 
        \|\matG^\top \vz\|_\infty
      \big), \\[0.3em]
\|\vr_e\|_\infty 
  &\leq \epsilon_a 
    + \epsilon_r \max\!\big(
        \|\matA \vx\|_\infty, 
        \|\vecB\|_\infty
      \big), \\[0.3em]
\|\vr_i\|_\infty 
  &\leq \epsilon_a 
    + \epsilon_r \max\!\big(
        \|\matG \vx\|_\infty, 
        \|\vs\|_\infty, 
        \|\vecH\|_\infty
      \big), \\[0.3em]
\|\vr_n\|_\infty 
  &\leq \epsilon_a 
    + \epsilon_r \max\!\big(
        \|\vs\|_\infty, 
        \|\vz\|_\infty, 
        \|\vr_g\|_\infty
      \big),
\end{aligned}
\end{equation}
where $\vr_n = \min(\vs,\vz)$ denotes the minimum function, and $\epsilon_a > 0$ and $\epsilon_r \ge 0$ are the absolute and relative tolerances used in the stopping criterion. 
We use the problem residuals $\vr_d$, $\vr_e$, and $\vr_i$, together with $\vr_g$, to scale the \gls{ncp}-based feasibility measure.

\subsection{Infeasibility detection}
Primal and dual infeasibility are certified using Farkas-type conditions, following the approaches in~\cite{banjac2019infeasibility, liao2019fbstab}. 
In \textsc{Odyn}, a \gls{qp} is declared dual infeasible when the following termination criterion is satisfied:
\begin{align}
\|\matQ\dx\|_{\infty} &\le \epsilon_d \|\dx\|_{\infty}, &
\vecC^{\transpose}\vx &< -\epsilon_d, \nonumber\\
\|\matA\dx\|_{\infty} &\le \epsilon_d\|\dx\|_{\infty}, &
\|\matG\dx + \ds\|_{\infty} &\le \epsilon_d \|\dx\|_{\infty},\label{eq:dual_certification}
\end{align}
where $\epsilon_p > 0$ is a dual infeasibility certification tolerance.
Moreover, the primal infeasible is certified when the following condition is met:
\begin{equation}\nonumber
\|\matA^\transpose\dy + \matG^\transpose\dz\|_\infty \leq \epsilon_{p}(\|\dy\|_\infty + \|\dz\|_\infty),
\end{equation}
\begin{equation}
\vecB^{\transpose}\dy + \vecH^{\transpose}\dz < -\epsilon_p,
\end{equation}
where $\epsilon_p > 0$ is a primal infeasibility certification tolerance. 

\subsection{Overall algorithm and software}
The overall algorithm of \textsc{Odyn}~\gls{qp} solver is presented in~\Cref{alg:odyn}.
\begin{algorithm}[b]
    \tcc{\textsc{Odyn} \gls{qp} iterations}
    \For{$i \leftarrow 0$ \KwTo $max\_iters$}{\label{alg:odyn_loop}
        search direction: $\Delta\U$ \hfill\Cref{eq:odyn_search_direction}\\
        flexible Armijo line search: $\alpha$ \hfill\Cref{proposed_armijo_equations}\\
        \If{$\|\tilde{\vr}(\U^{+\alpha};\boldsymbol{\Pi})\|_\infty 
            \leq \theta \|\tilde{\vr}(\U;\boldsymbol{\Pi})\|_\infty + \beta\mu$}{
            update centering \hfill\Cref{alg:centering_algorithm}\\
            update step-length estimates \hfill\Cref{alg_compute_step_length_estimates}\\
            update proximal primal--dual estimates \hfill\Cref{eq_proximal_primal_dual_estimates}
        }
        test stopping criteria \hfill\Cref{eq:stopping_criteria}
    }
    \caption{\textsc{Odyn}~\gls{qp} solver}
    \label{alg:odyn}
\end{algorithm}
\textsc{Odyn} is distributed as an open-source implementation.\footnote{The code will be available after acceptance.}
The solver is written in \textsc{C++} with \textsc{Python} bindings and relies on the \textsc{Eigen} library~\cite{eigen} for efficient linear-algebra operations in both its dense and sparse backends.

Formulating and solving a \gls{qp} with \textsc{Odyn} involves four main components: \texttt{Model}, \texttt{Data}, \texttt{Params}, and \texttt{Solver}. 
All components, except for the \texttt{Params} structure, are specialized for both dense and sparse backends. 
Below, we illustrate how to set up and solve \gls{qp} problems for each backend in \textsc{Python}.
Additionally, the default hyperparameters employed by \textsc{Odyn} (using double precision numbers) are outlined in~\Cref{tab:default_hyperparams}.
For single precision numbers, we increase $\boldsymbol{\rho}$ and $\mu_{\min}$ by $10^4$.
Next, we introduce applications of \textsc{Odyn} in the contexts of predictive control, deep learning, and simulation.

\begin{lstlisting}[caption={Example illustrating \textsc{Odyn}'s API.}, label={lst:odyn}]
import odyn

# Formulate and solve a dense QP problem.
# We start creating a random QP model, data, as well as a QP solver and parameters.
# The 'print_level' parameter controls the verbosity of the output, with 'odyn.VerboseLevel.High' being the most verbose.
model = odyn.DenseModel.Random(10, 5, 3)
data = model.createData()
params = odyn.Params()
solver = odyn.DenseQP()
solver.solve(model, data, params, print_level=odyn.VerboseLevel.High)

# If we want to run the sparse backend for this random QP problem, we can just convert the model and solver to sparse format.
sparse_model = model.toSparse()
sparse_data = sparse_model.createData()
sparse_solver = odyn.SparseQP()
sparse_solver.solve(
    sparse_model, sparse_data, params, print_level=odyn.VerboseLevel.High)

# We can create a dense models from the QP's matrices and vectors
other_model = odyn.DenseModel(model.Q, model.c, model.A, model.b, model.G, model.h)
\end{lstlisting}

\begin{table}[b]
  \centering
  \caption{\textsc{Odyn}'s hyper-parameters for double precision.}
  \label{tab:default_hyperparams}
  \renewcommand{\arraystretch}{1.2}
  \begin{tabular}{lcl}
    \hline
    \textbf{Parameter} & \textbf{Value} & \textbf{Description} \\
    \hline
    $\beta$, $\theta$          & $0.85, 0.95$      & neighborhood hyper-parameters  \\
    $\sigma_\text{min}$, $\sigma_\text{max}$ & $0.1, 0.9$ & minimum and maximum centering \\
    $\mu_\text{min}$            & $10^{-16}$   & minimum barrier value \\
    $\theta_\text{min}$       & $0.1$   & minimum interpolation value \\
    $\theta_l$, $\theta_u$     & $0.3, 0.85$   & min. and max. improvement thresholds \\
    $\theta^{-}$, $\theta^{+}$ & $0.4, 0.3$   & decrease and increase constants \\
    $\boldsymbol{\rho}$                   & $10^{-9}$   & proximal penalty parameters\\
    $\rho_\text{min}$        & $10^{-9}$   & minimum penalty parameters\\
    $\delta$        & $5$   & regularization decreasing factor \\
    $\eta$          & $10^{-2}$ & line search parameter\\   
    \hline
  \end{tabular}
\end{table}

\section{\textsc{Odyn} Application in Robotics and AI}
In this section, we briefly introduce three applications of \textsc{Odyn}:  
(i) an \gls{sqp} solver built on top of the \textsc{Odyn} \gls{qp} backend (\texttt{OdynSQP}),  
(ii) a \textsc{PyTorch} module that embeds \textsc{Odyn} as an implicitly differentiable layer (\texttt{ODYNLayer}), and  
(iii) a contact-physics engine that uses \textsc{Odyn} as its numerical backend (\texttt{ODYNSim}).  
The following subsections provide essential details on how \textsc{Odyn} is integrated into each of these components using state-of-the-art techniques.
Further implementation details of these applications lie beyond the scope of this article and will be released as part of the open-source implementation.

\subsection{\texttt{OdynSQP}: A~\gls{sqp} solver for~\gls{mpc}}
\gls{qp} solvers are essential building blocks for solving nonlinear programs through~\acrfull{sqp}.
The core idea of~\gls{sqp} is to transform a nonlinear program into a sequence of~\glspl{qp}, each solving a local linear-quadratic approximation of the cost and constraints.

\gls{sqp} methods are among the most powerful tools for nonlinear optimal control and estimation in robotics.
Specifically, they can be applied to solve optimal control problems defined as follows
\begin{equation}\label{eq:oc_formulation}
\begin{aligned}
\min_{\mathbf X, \mathbf U} \quad 
    & \ell_N(\state_N) + \sum_{k=0}^{N-1} \ell_k(\state_k, \control_k) \\
\text{subject to} \quad 
    & \state_0 = \hat{\state}_0, \\ 
    & \state_{k+1} = \dynFun_k(\state_k, \control_k), \\ 
    & \eqFun_k(\state_k, \control_k) = \mathbf 0, \\ 
    & \ineqFun_k(\state_k, \control_k) \le \mathbf 0,
\end{aligned}
\end{equation}
where $\state=(\pos,\vel)\in\stateManif\subseteq\R^\nx$ denotes the system state, $\control\in\R^\ntau$ the control input, $\ell_N(\state_N)$ the terminal cost, and $\ell_k(\state_k, \control_k)$ is the stage cost at node $k$.
The system evolution is governed by the dynamics $\dynFun:\stateManif\times\R^\ntau$, while $\eqFun:\stateManif\times\control$ and $\ineqFun:\stateManif\times\control$ encode equality and inequality path constraints.
In robotics, equality constraints commonly enforce end-effector constraints~\cite{giftthaler-ichr17,parilli-endpointmpc} or embed inverse dynamics~\cite{mastalli-invdynmpc,ferrolho-icra21}, whereas inequality constraints capture joint limits (e.g.,~\cite{mastalli22auro}), friction cones, physical realism of dynamics parameters (e.g.,~\cite{martinez2025multi}), and other safety or feasibility conditions (e.g.,~\cite{caron-icra15,ordonez2024morphological}).
This sequence of~\gls{qp} problems are sparse with block-banded structure.

At each iteration,~\gls{sqp} forms a quadratic approximation of the Lagrangian, yielding the following~\gls{qp}:
\begin{equation}\label{eq:sqp_formulation}
\begin{aligned}
\min_{\xu} \quad 
    & \frac{1}{2}\delta\xu\,\mathbf{H}(\xu_i)\,\delta\xu + \nabla\mathbf{J}(\xu_i)^\transpose\delta\xu \\
\text{subject to} \quad 
    & \bar{\eqFun}(\xu_i) + \nabla\bar{\eqFun}(\xu_i)\delta\xu = \bzero, \\
    & \bar{\ineqFun}(\xu_i) + \nabla\bar{\ineqFun}(\xu_i)\delta\xu \leq \bzero,
\end{aligned}
\end{equation}
where $\xu$ stacks the state trajectory and control sequence.
The matrices $\mathbf{H}(\xu_i)$ and $\nabla\mathbf{J}(\xu_i)$ are assembled from the derivatives of the stage and terminal costs at the current iterate $\xu_i$, i.e., $\ell_\state$, $\ell_\control$, $\ell_{\state\state}$, $\ell_{\state\control}$, and $\ell_{\control\control}$. 
Similarly, $\nabla\bar{\eqFun}(\xu_i)$ is obtained by linearizing the dynamics ($\dynFun_\state$, $\dynFun_\control$) and stagewise equality constraints ($\eqFun_\state$, $\eqFun_\control$), and $\nabla\bar{\ineqFun}(\xu_i)$ is built from the linearized stagewise inequality constraints ($\ineqFun_\state$, $\ineqFun_\control$).

\texttt{OdynSQP} is implemented using \textsc{Crocoddyl}~\cite{mastalli2020} and \textsc{Odyn}'s sparse backend. 
In addition, we incorporate a Levenberg--Marquardt scheme to enhance convergence robustness, together with a non-monotone Armijo-type line search for step acceptance.

\subsection{\texttt{ODYNLayer}: Differentiable optimization}
Differentiable optimization~\cite{optnet,bambade2024qplayer,magoon_DiffQP} has gained attention in differentiable physics simulation~\cite{newton, gradsim} and policy learning~\cite{xu2021accelerated}.
This is because it enables gradient-based training through implicit optimization problems.

Embedding a differentiable~\gls{qp} layer within a neural network allows joint optimization of perception and control parameters while preserving structural priors such as dynamics, constraints, and energy conservation.
This approach underpins many \emph{model-based reinforcement learning} methods~\cite{MBRL}, where differentiable rollouts integrate control subproblems (e.g.,~\gls{mpc} or contact physics), yielding task-consistent gradients through structured decision processes instead of unstructured policy gradients.

\subsubsection{Parametrized~\glspl{qp}}
A parameterized quadratic program $\textrm{QP}(\boldsymbol{\theta})$ defines a family of~\gls{qp} solutions depending on parameters $\boldsymbol \theta$:
\begin{equation}\label{eq:param_qp}
\begin{aligned}
\min\limits_{\vx, \vs} \quad
    & \tfrac{1}{2}\,\vx^\transpose \matQ(\boldsymbol{\theta})\,\vx
      + \vecC(\boldsymbol{\theta})^\transpose \vx \\
\text{subject to} \quad
    & \matA(\boldsymbol{\theta})\,\vx = \vecB(\boldsymbol{\theta}), \\
    & \matG(\boldsymbol{\theta})\,\vx + \vs = \vecH(\boldsymbol{\theta}), \\
    & \vs \ge \bzero,
\end{aligned}
\end{equation}
where its corresponding~\gls{kkt} point $\U^\star = (\vx^\star, \vy^\star, \vs^\star, \vz^\star)$ depend on the parameters $\boldsymbol \theta$ and its computed by satisfying the perturbed~\gls{kkt} residual $\tilde{\vr}(\U^\star;\boldsymbol{\Pi}) = \bzero$.
For a scalar objective $\ell(\U^\star, \boldsymbol{\theta})$, the derivative is computed implicitly through the optimality conditions rather than by unrolling the solver’s iterative steps. 
This is enabled by the \gls{ift} (see~\cite{IFT}).

The resulting approach is both numerically stable and computationally efficient, as it avoids the long backpropagation chains that often cause exploding or vanishing gradients when differentiating through iterative optimization procedures~\cite{optnet}.

\subsubsection{Gradients via~\gls{ift}}
Linearizing $\tilde{\vr}(\U^\star;\boldsymbol{\Pi}, \boldsymbol{\theta})$ at $\U^\star$ yields
\begin{equation}
\left .\frac{\partial \tilde{\vr}}{\partial{\U}}\right|_{\U=\U^{\star}} \mathrm{d} \U^\star + \frac{\partial \tilde{\vr}}{\partial \boldsymbol{\theta}}\, \mathrm{d}\boldsymbol{\theta} = \bzero.
\end{equation}
According to the~\gls{ift}, \emph{iff} $\partial \tilde{\vr}/\partial \U$ is non-singular,
\begin{equation}\label{eq:odynlayer-ift}
\begin{aligned}
\left.\frac{\partial \U^\star}{\partial \boldsymbol{\theta}}\right|_{\U=\U^{\star}}
&= 
-\left[\left.\frac{\partial \tilde{\vr}}{\partial \U}\right|_{\U=\U^{\star}}\right]^{-1}
\frac{\partial \tilde{\vr}^{\star}}{\partial \boldsymbol{\theta}}, \\[0.6em]
\text{with}\quad
\frac{\partial \tilde{\vr}^{\star}}{\partial \boldsymbol{\theta}}
&=
\begin{bmatrix}
    \dfrac{\mathrm{d}\mathbf Q}{\mathrm{d}\boldsymbol{\theta}}\,\mathbf x^{\star}
    + \dfrac{\mathrm{d}\mathbf c}{\mathrm{d}\boldsymbol{\theta}}
    + \left(\dfrac{\mathrm{d}\mathbf A}{\mathrm{d}\boldsymbol{\theta}}\right)^{\transpose}\mathbf y^{\star}
    + \left(\dfrac{\mathrm{d}\mathbf G}{\mathrm{d}\boldsymbol{\theta}}\right)^{\transpose}\mathbf z^{\star} \\
    \dfrac{\mathrm{d}\mathbf A}{\mathrm{d}\boldsymbol{\theta}}\,\mathbf x^{\star}
    - \dfrac{\mathrm{d}\mathbf b}{\mathrm{d}\boldsymbol{\theta}} \\
    \dfrac{\mathrm{d}\mathbf G}{\mathrm{d}\boldsymbol{\theta}}\,\mathbf x^{\star}
    - \dfrac{\mathrm{d}\mathbf h}{\mathrm{d}\boldsymbol{\theta}} \\
    \mathbf 0
\end{bmatrix}.
\end{aligned}
\end{equation}

To avoid explicit Jacobian calculations, we introduce the adjoint $\boldsymbol \lambda$ as the solution of the linear system,
\begin{equation}\label{eq:odynlayer_adjoint}
\left.\left[\frac{\partial \tilde{\vr}}{\partial \U}\right|_{\U = \U^{\star}}\right]^{\transpose}
\boldsymbol{\lambda}
= 
-\left.\frac{\partial \ell}{\partial \U}\right|_{\U = \U^{\star}},
\end{equation}
while the hyper-gradient is defined using the solution $\boldsymbol \lambda^{\star}$,
\begin{equation}\label{eq:adj_eq2}
\frac{\partial \ell}{\partial \boldsymbol{\theta}} = (\boldsymbol \lambda^{*})^\transpose\frac{\partial \tilde{\vr}}{\partial \boldsymbol{\theta}}.
\end{equation}
Crucially, $\partial \tilde{\vr} / \partial \U$ coincides with the~\gls{kkt} matrix used in the forward Newton step (\Cref{eq:full_kkt_system,eq:reduced_kkt_system}), allowing the backward pass to efficiently compute the  gradients of feasible~\glspl{qp}.
In cases where the~\gls{kkt} matrix becomes ill-conditioned or nearly singular at convergence, a regularized variant of~\Cref{eq:adj_eq2} is solved to ensure numerical stability.

\texttt{ODYNLayer} is integrated into \textsc{PyTorch}.
It supports both dense and sparse backends, multiple floating-point precisions, as well as condensed and full \gls{kkt} factorizations.

\subsection{\texttt{ODYNSim}: contact dynamics as~\gls{qp}}
Among all forces admissible under the unilateral and friction constraints, the realised contact force is the one that maximises energy dissipation.
This dissipative nature of frictional contact is stated by maximum dissipation principle~\cite{moreau-contactdynbook}. 
For rigid contacts with relative velocity $\vel$, this leads directly to a convex conic program.
However, if we approximate the Coulomb friction cone for dry friction with a pyramidal model, the problem can be cast as a \gls{qp} of the form
\begin{equation}\label{eq:contact_qp}
\begin{aligned}
\min\limits_{\force} & \quad \frac{1}{2}\force^\transpose \delassusMat(\pos)^{-1}\force + \force^\transpose ( \freeContVel(\pos)^- + \deSaxce(\vel))\\ 
\text{subject to} & \quad \mathbf{C}\force \geq \bzero,
\end{aligned}
\end{equation}
where $\force\in\R^{3\nc}$ describes the contact-force vector, $\delassusMat(\pos)=\contactJac(\pos)\massMat(\pos)^{-1}\contactJac(\pos)^\transpose\in\R^{3\nc\times3\nc}$ is the Delassus matrix with $\massMat(\pos)\in\R^{\nq\times\nq}\succeq\bzero$ and $\contactJac(\pos)\in\R^{3\nc\times\nq}$ as the joint-space mass matrix and contact Jacobian, respectively, $\freeContVel(\pos)^- =\contactJac(\pos)\vel^-\in\R^{3\nc}$ is the free velocity of the contact, $\deSaxce(\vel)\in\R^{3\nc}$ is the de-Saxcé correction~\cite{desaxce-bipotential}, and $\mathbf{C}$ defines a pyramidal approximation of the Coulomb friction cone.

\section{Results}
We begin by benchmarking the convergence behavior of our \gls{qp} solver against state-of-the-art methods developed over the past decade using the Maros--Mészáros test set (\Cref{sec:hard_qp_results}). 
We then compare the warm-starting capabilities of \textsc{Odyn} with those of leading \gls{alm}-based solvers (\Cref{sec:warmstarting_results}). 
Next, we examine the behavior of \textsc{Odyn} on a set of representative degenerate problems (\Cref{sec:degenerate_results}). 
Finally, we report results for \texttt{OdynSQP}, \texttt{ODYNSim}, and \texttt{ODYNLayer} in \Cref{sec:odynsqp_results,sec:odynsim_results,sec:odynlayer_results}. 
Together, these experiments demonstrate the suitability of \textsc{Odyn} for robotics and AI applications. 
All experiments were conducted on a MacBook Pro equipped with an M2~Pro processor.

\subsection{Hard \gls{qp} problems: the Maros--Mészáros test set}\label{sec:hard_qp_results}
The Maros--Mészáros test set introduced in~\cite{maros1999repository} comprises $138$~\gls{qp} problems and serves as a standard benchmark for evaluating the performance of~\gls{qp} solvers.
They are widely regarded as the gold standard for benchmarking \gls{qp} solvers.

Using the  Maros--Mészáros test set, we evaluated \textsc{Odyn} against state-of-the-art sparse~\gls{qp} solvers, including \textsc{ProxQP}, \textsc{PiQP}, \textsc{OsQP}, \textsc{Mosek}, and \textsc{Gurobi}, under both medium- and high-accuracy configurations.
We reported statistic of performance such as failure rate and performance profiles with absolute and relative tolerances set to $10^{-6}$ or $10^{-9}$, i.e., medium and high accuracy, respectively. 

\subsubsection{Failure rates}
\Cref{tab:failure_rate} summarizes the failure rates under both medium- and high-accuracy settings. 
The results indicate that \textsc{Odyn} is among the most competitive solvers and achieves the best performance at medium accuracy.
Its strong convergence robustness can be attributed to its close connection with interior-point methods. 
Notably, \glspl{ipm} (\textsc{PiQP}, \textsc{MOSEK}, and \textsc{Gurobi}) consistently outperform \glspl{alm} (\textsc{ProxQP} and \textsc{OsQP}) in this benchmark.
Moreover, the slower convergence observed in \textsc{Odyn} under high-accuracy settings may be attributed to its lack of iterative refinement procedures, which are commonly employed in \gls{qp} solvers.
Note that most failures observed with \textsc{Odyn} were due either to exceeding the time limit ($120$ seconds) or, in some cases, to the detection of primal or dual infeasibility.

\begin{table}[t]
\centering
\caption{Failure rate (\%) of different~\gls{qp} solvers on a subset of the Maros--Mészáros test set at medium and high accuracy.}
\label{tab:failure_rate}
\begin{tabular}{@{}l*{6}{c}@{}}
\toprule
Accuracy  & \textsc{Odyn} & \textsc{ProxQP} & \textsc{PiQP} 
          & \textsc{OsQP} & \textsc{Mosek} & \textsc{Gurobi} \\
\midrule
Medium    &  \textbf{9.7\%} & 33.0\% & 10.7\% & 44.7\% & 14.6\% &  \textbf{9.7\%} \\
High      & 15.5\% & 35.9\% & \textbf{10.7\%} & 49.5\% & 14.6\% & 12.6\% \\
\bottomrule
\end{tabular}
\end{table}


\subsubsection{Performance profiles}\label{sec:performance_profiles}
\begin{figure}[!t]
\centering
\includegraphics[width=1\linewidth]{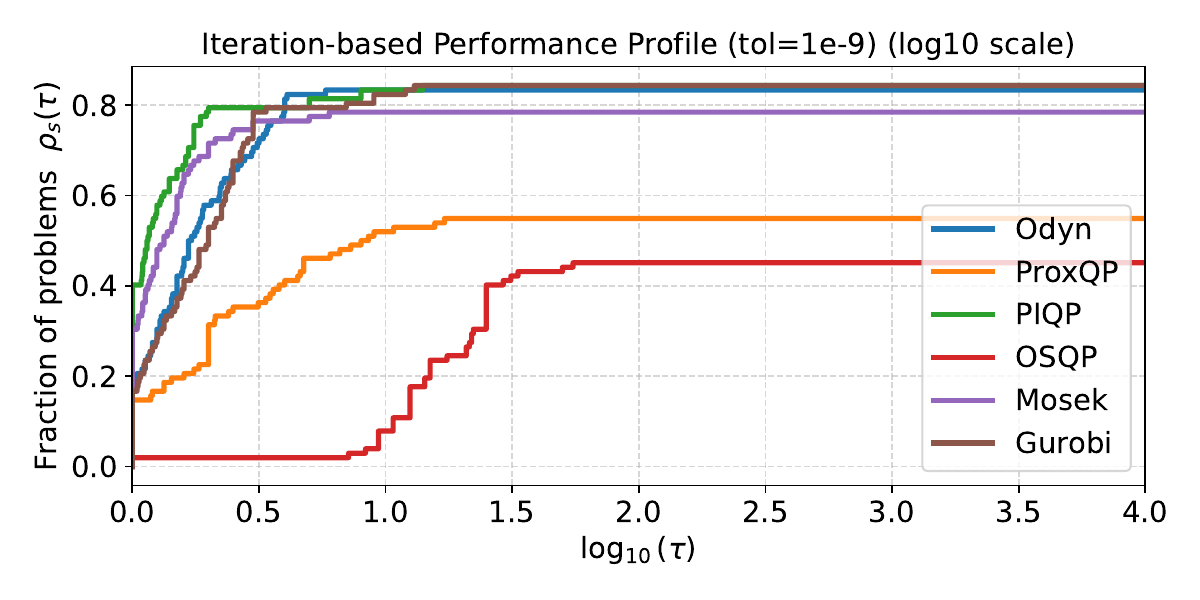}
\includegraphics[width=1\linewidth]{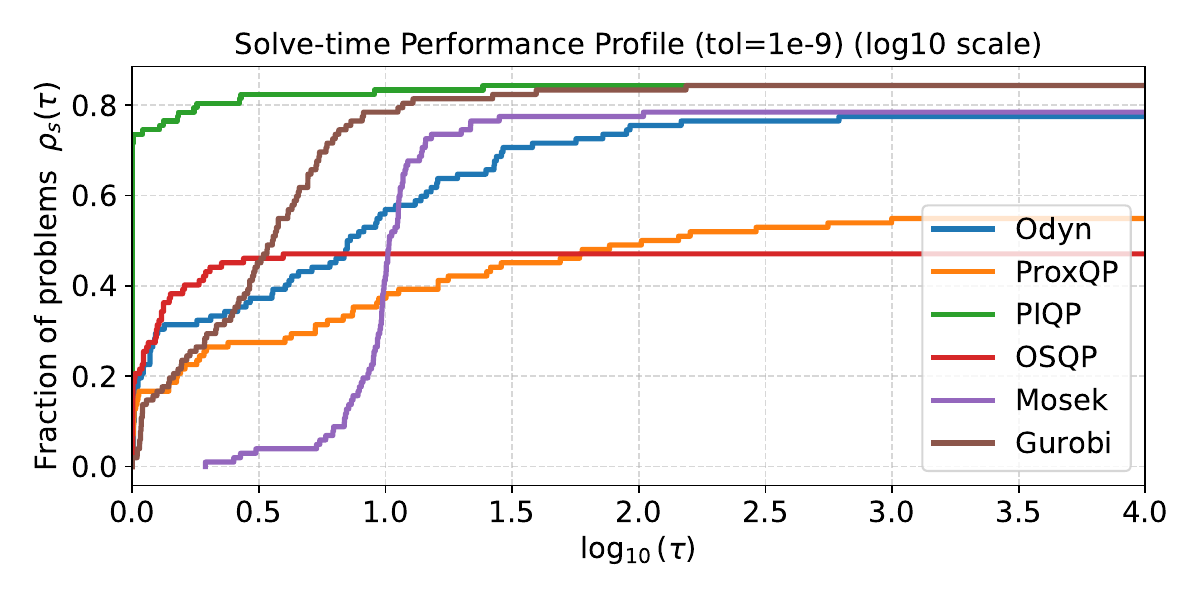}
\caption{(top) Iteration based performance profiles at high accuracy.
         (bottom) Solve time performance profiles at high accuracy.}
\label{fig:performance_plots}
\end{figure}

We employed performance profiles~\cite{dolan2002benchmarking} to evaluate and compare \textsc{Odyn} with state-of-the-art solvers. 
Profiles based on computation time and iteration count reported in~\Cref{fig:performance_plots} were obtained by running a set of solvers $\mathcal{S}$ on a collection of benchmark problems~$\mathcal{P}$.
Each performance profile reports the fraction of problems solved by a specific solver, which is expressed as a function of the best observed performance for a given metric.
Specifically, we defined the performance ratio of solver $s \in \mathcal{S}$ on problem $p \in \mathcal{P}$ as:
\begin{equation}
r_{p,s} = \frac{t_{p,s}}{\min_{s\in\mathcal{S}} t_{p,s}},
\end{equation}
where $t_{p,s}$ represents computation time or number of iterations (our metrics).
When a solver $s$ failed to solve problem $p$,  we set $t_{p,s}$ to an upper bound like the maximum allowed time to solve a problem or a maximum number of iterations.

The performance of a solver $s$ on a test set of problems $\mathcal{P}$ was defined as:
\begin{equation}
f_s(\tau) = \frac{1}{|\mathcal{P}|}\sum_{p\in\mathcal{P}}\mathcal{I}_{\leq\tau}(r_{p,s}),
\end{equation}
where $\mathcal{I}_{\leq\tau}(r_{p,s}) = 1$ if $r_{p,s}\leq\tau$ or $0$ otherwise.

\subsection{Warm-starting performance comparison}\label{sec:warmstarting_results}
We evaluated the warm-starting capabilities of \textsc{Odyn} against \textsc{OsQP} and \textsc{ProxQP} on a subset of the Maros--Mészáros test that are feasible and solvable by all the solvers.
When doing so, we used the high-accuracy setting.
Note that \textsc{PiQP}, \textsc{Mosek}, and \textsc{Gurobi} were excluded as interior-point solvers do not support warm-starting.

We assessed the warm-starting capabilities by adopting an approach commonly used in the optimization (see~\cite{benson2007}).
Concretely, we first solved the original problem from a cold-start initialization using the~\gls{qp} problem $\mathcal{QP} = (\matQ, \vecC, \matA, \vecB, \matG, \vecH)$.
With this primal--dual solution, we recorded the number of iterations required for convergence, denoted by $N_{\text{cold}}$.
We then constructed a perturbed problem $\tilde{\mathcal{QP}} = (\Qtil, \tilde{\vecC}, \tilde{\matA}, \tilde{\vecB}, \tilde{\matG}, \tilde{\vecH})$ as described in~\cite{engau2009primal}.
For instance, the perturbed right-hand side $\tilde{\vecB}$ is computed as $\tilde{\vecB} = \vecB + \delta(\vecM \circ \vecEta \circ \vecB)$, where $\delta>0$ describes the different perturbation levels $(0.001,0.01,0.1)$, $\vecEta \sim \mathrm{Uniform}([-1,1])^m$ is a uniformly distributed random vector.
Moreover, $\vecM$ is a random mask vector defined as:
\begin{equation}
\vecM_i =
\begin{cases}
1, & \text{if } \vecZeta_i < \min(0.1,\; 20/m)\\
0, & \text{otherwise}
\end{cases},
\end{equation}
where $\vecZeta \sim \mathrm{Uniform}([-1,1])^m$ is another uniformly distributed random vector.
Again, this strategy was used to perturb the entire~\gls{qp} problem.
However, when doing so, we preserved the same sparsity structure and symmetric of the original matrices $\matQ$, $\matA$, and $\matG$.

\begin{figure}[!t]
\centering
\includegraphics[width=1\linewidth]{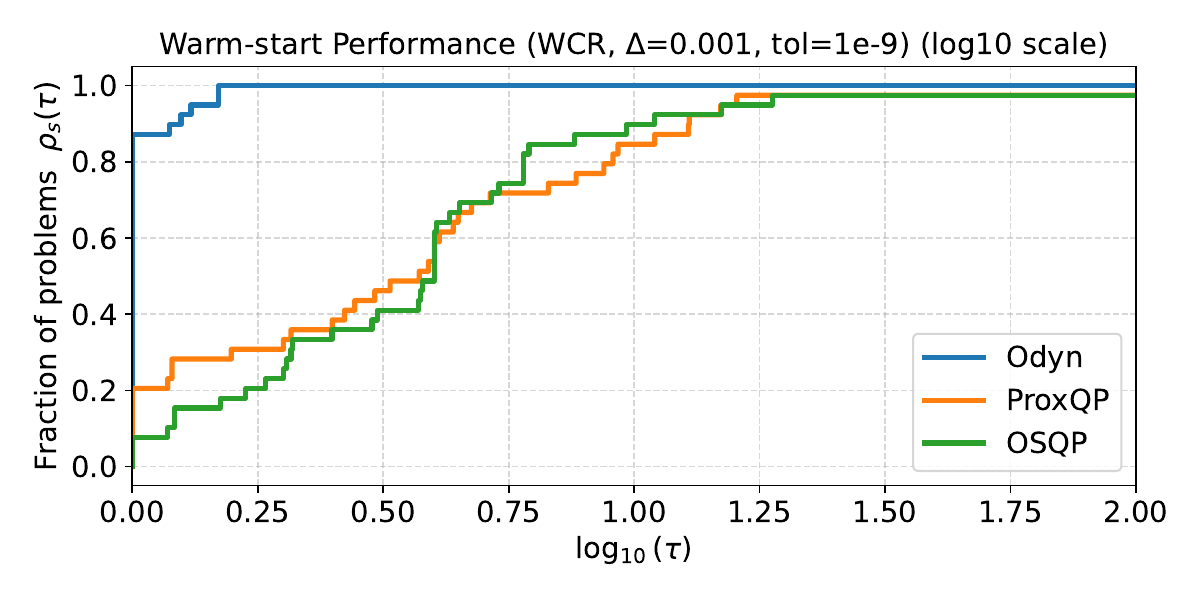}
\includegraphics[width=1\linewidth]{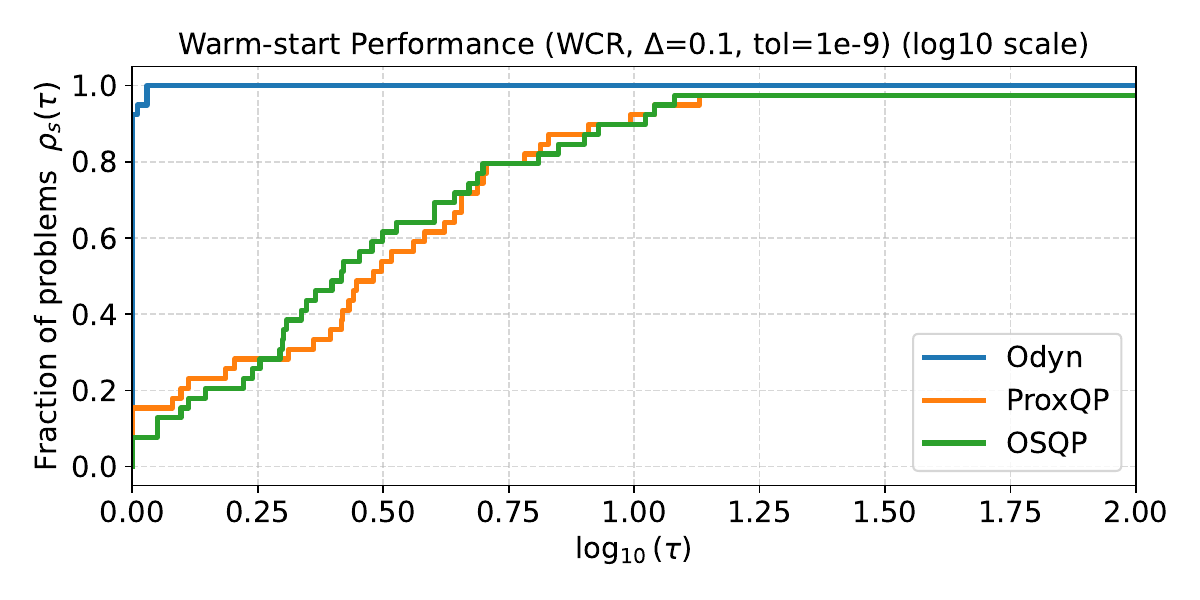}
\caption{Warm-to-cold ratio performance profiles under high-accuracy settings, evaluated with perturbed \gls{qp} data.
The top and bottom plots correspond to perturbation levels $\delta = 10^{-3}$ and $10^{-1}$, respectively.
}
\label{fig:warm_high}
\end{figure}

For each perturbation level $\delta$, we recorded the number of iterations required to solve $10$ perturbed models $\tilde{\mathcal{QP}}_i$, with $i \in \{1, \ldots, 10\}$. 
This allowed us to compute the mean \emph{warm-to-cold ratio}, defined as $\text{WCR} = N_{\text{warm}} / N_{\text{cold}}$, which quantifies the reduction in iteration count achieved through warm-starting relative to a cold start. 
\Cref{fig:warm_high} reports the warm-starting performance across different perturbation levels, showing~\textsc{Odyn}'s superiority compared to \textsc{OsQP} and \textsc{ProxQP}.
These results are reported for high-accuracy settings; however, a similar behavior was observed for medium accuracy.
Additionally, these plots were generated using the standard benchmarking procedure described in~\Cref{sec:performance_profiles}, which is more rigorous than directly plotting the WCR against the fraction of problems below a fixed threshold, as done in~\cite{engau2009primal}.

\subsection{Dense backend}
\begin{figure}[!t]
\centering
\includegraphics[width=1\linewidth]{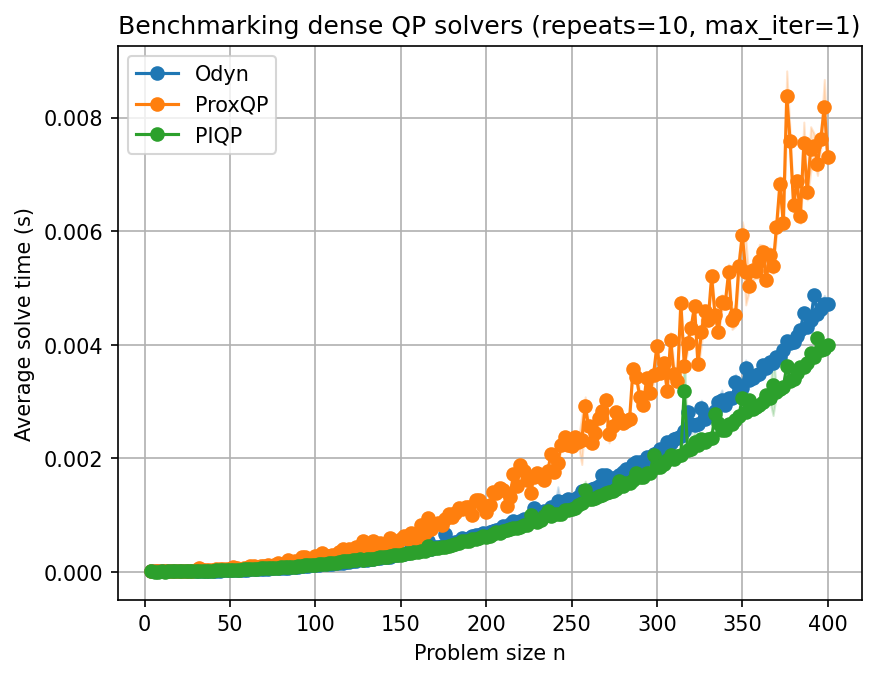}
\caption{Performance of the dense backend of \textsc{Odyn}, \textsc{ProxQP}, and \textsc{PiQP}.
We generate random~\gls{qp} problems with dimensions ranging from $1$ to $400$ and report the average computation time per iteration over $10$ trials.
As expected, all solvers exhibit cubic complexity, whereas \textsc{Odyn} and \textsc{PiQP} demonstrate more deterministic timing behaviour.
}
\label{fig:dense-perf-per-size}
\end{figure}

We also benchmarked the dense backends of \textsc{Odyn}, \textsc{ProxQP}, and \textsc{PiQP}.
\Cref{fig:dense-perf-per-size} shows the average per-iteration computation time versus problem size.
For each dimension, we generated random~\gls{qp} instances with $m = p =\frac{n}{2}$ and repeated the solve $10$ times.
It can be observed that \textsc{Odyn} is competitive.
These results are critical for many real-time applications in robotics, including model predictive control, whole-body control, and contact simulation.
They are also relevant for training differentiable \gls{qp} layers in AI, where many applications are expected to involve small- to medium-scale, dense problem instances.

\subsection{Degenerate~\gls{qp} problems}\label{sec:degenerate_results}
We analyzed the robustness of \textsc{Odyn} on a set of degenerate problems. 
Iteration counts were benchmarked by treating all problems as sparse. 
The results, reported in~\Cref{tab:degenerate_qp_iters}, show that \textsc{Odyn} is competitive with other \gls{qp} solvers.
It also indicates that \textsc{OsQP}, a widely used solver in robotics due to its warm-starting capabilities, exhibits reduced numerical stability under the tested conditions.
We selected simple \gls{qp} problems so that they also serve an educational purpose. 
The corresponding test cases are described below.

\begin{table}[t]
  \centering
  \caption{Number of iterations required by each solver on the three degenerate~\gls{qp} problems.
  We highlight the three best solvers.}
  \label{tab:degenerate_qp_iters}
  \renewcommand{\arraystretch}{1.15}
  \setlength{\tabcolsep}{4pt} 
  \begin{tabular}{lccc}
    \hline
    \textbf{Solver} &
    \textbf{Redundant Ineq.} &
    \textbf{\acrshort{licq} Failure} &
    \textbf{Multi. Solutions} \\
    \hline
    \textsc{Odyn}   & \textbf{5}   & \textbf{10}  & \textbf{6}   \\
    \textsc{ProxQP} & 12  & 30  & 10  \\
    \textsc{PiQP}   & \textbf{5}   & \textbf{8}   & \textbf{6}   \\
    \textsc{OsQP}   & 75  & $\dagger$ & 200 \\
    \textsc{Mosek}  & \textbf{4}   & 14  & \textbf{5}   \\
    \textsc{Gurobi} & 11  & \textbf{8}   & 9   \\
    \hline
  \end{tabular}
  \vspace{2pt}
  {\\\footnotesize
  \textsuperscript{$\dagger$}\textsc{OsQP} declared \emph{dual infeasible} on the~\acrshort{licq} failure problem.
  }
\end{table}

\subsubsection{Redundant constraints}
We propose a simple \gls{qp} that exhibits ill-conditioning due to redundant inequality constraints as follows
\begin{equation}\label{eq:degenerate_qp1}
\begin{aligned}
\min\limits_{x_1, x_2} \quad 
    & \frac{1}{2}\left(x_1^2 + x_2^2\right) \\
\text{subject to} \quad 
    & x_1 \geq 1, \\
    & x_1 \geq 0.
\end{aligned}
\end{equation}
In this~\gls{qp} problem, we observe that the second inequality constraint is redundant, yet it may still introduce numerical instability in many~\gls{qp} solvers.

\subsubsection{Linear dependent constraints}
We consider a \gls{qp} problem that fails to meet the~\gls{licq} at convergence, while being ill-conditioned as its $\matQ$ matrix and $\vecC$ vector are filled with small entries, i.e.,
\begin{equation}\label{eq:degenerate_qp2}
\begin{aligned}
\min\limits_{x_1, x_2} \quad 
    & \tfrac{1}{2}\!\left(10^{-10}x_1^2 + 10^{-6}x_2^2\right)
      \\ & \quad + 10^{-12}x_1 x_2 + 10^{-4}x_1 - x_2 \\
\text{subject to} \quad 
    & x_1 = 0, \\
    & x_1 \geq 0, \\
    & x_2 \geq 0.
\end{aligned}
\end{equation}

In this~\gls{qp} problem, the optimal solution is $\vx = \begin{bmatrix}0 & 0\end{bmatrix}$, requiring the activation of all the constraints.
This means that, when the inequality $x_1\geq 0$ becomes active at the solution, there is linearly dependency with the equality constraint, thereby violating~\gls{licq}.

\subsubsection{Multiple solutions}
We also study a \gls{qp} problem that admits multiple solutions due to a positive semidefinite $\matQ$ matrix, i.e.,
\begin{equation}\label{eq:degenerate_qp3}
\begin{aligned}
\min\limits_{x_1, x_2} \quad
    & \tfrac{1}{2}x_1^2 + x_1 \\
\text{subject to} \quad
    & 0 x_1 + 0 x_2 \leq 0, \\
    & 0 \leq x_1 \leq 3, \\
    & 0 \leq x_2 \leq 3.
\end{aligned}
\end{equation}

In this case, the variable $x_1$ is free to take any value in the interval $\begin{bmatrix}1,  3\end{bmatrix}$, and all such values yield the same optimal cost.
Additionally, the first constraint is a redundant inequality.
This is because it is always active but has no effect on the feasible set.
Moreover, its gradient is zero, which makes the~\gls{kkt} conditions ill-posed, since an active constraint with zero gradient violates the regularity conditions (e.g.,~\gls{licq}) typically required for well defined Lagrange multipliers.
Again, the results of solving these~\gls{qp} problems are summarized in~\Cref{tab:degenerate_qp_iters}.

\subsection{Optimizing agile maneuvers via \texttt{OdynSQP}}\label{sec:odynsqp_results}

\begin{figure*}[!t]
\centering
\includegraphics[width=\textwidth]{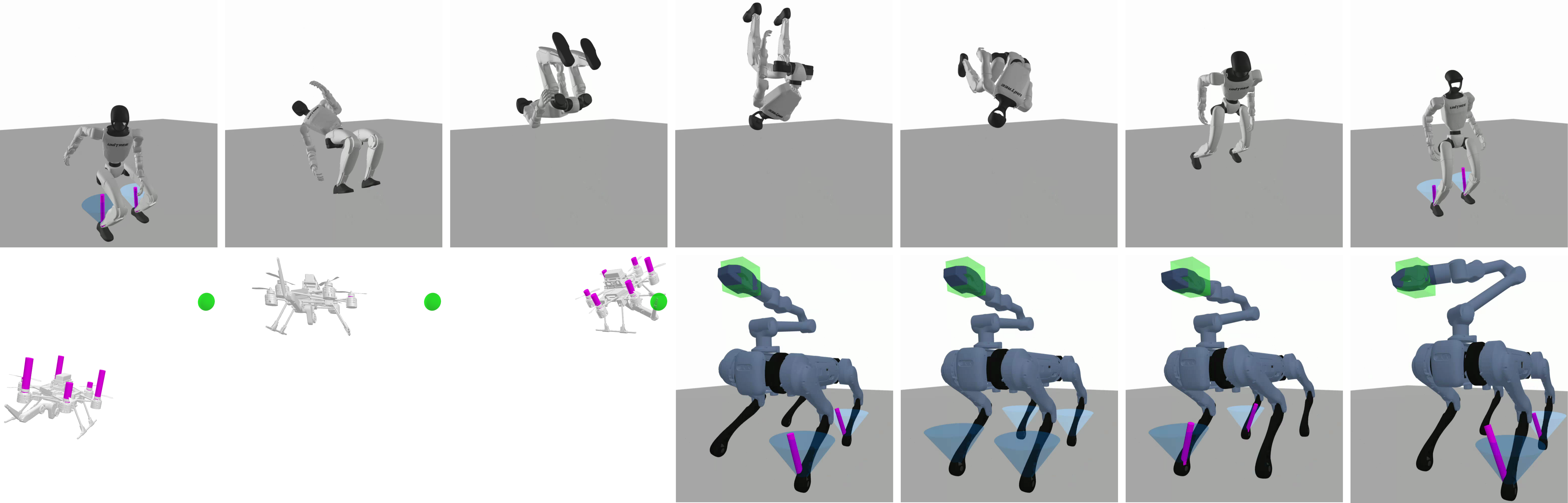}
\caption{Snapshots of optimal trajectories computed with \texttt{OdynSQP}. 
  (Top) G1 humanoid performing a backflip under joint position, velocity, and torque limits, together with contact wrench-cone constraints. 
  (Bottom-left) Borinot aerial robot reaching a target end-effector position (green sphere) while satisfying actuator torque limits. 
  (Bottom-right) B1--Z1 executing a chicken-head stabilization task during lateral locomotion while enforcing end-effector and friction-cone constraints; the robot’s hand is constrained to remain within the green bounding box. } 
  \label{fig:odynsqp_to_results}
\end{figure*}

\begin{figure}[!t]
\centering
\includegraphics[width=3.4in]{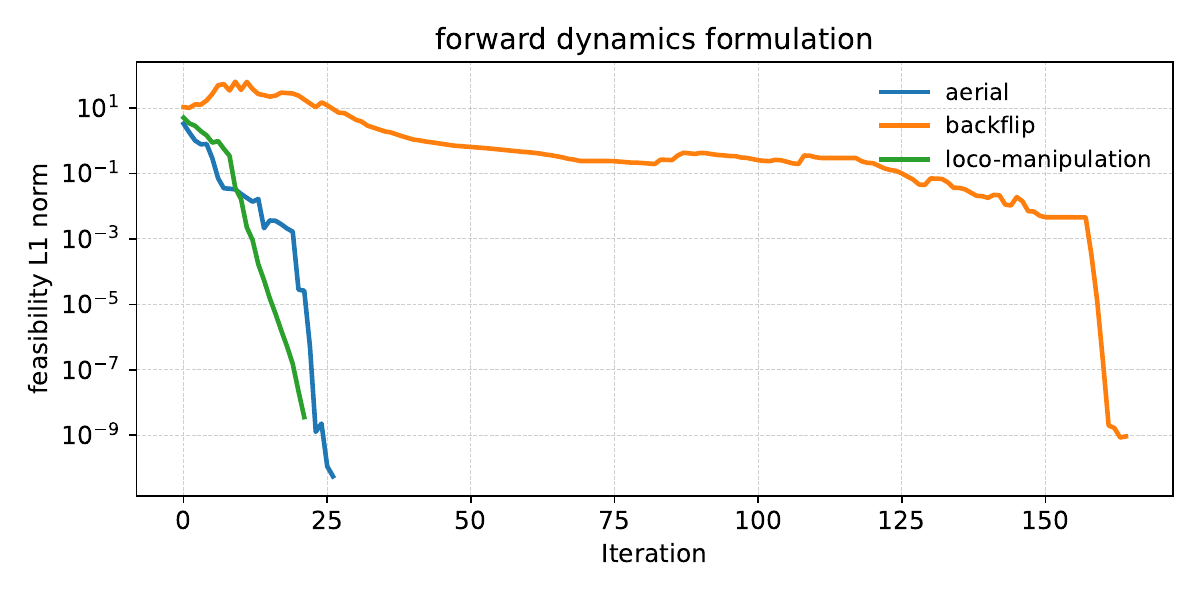}
\caption{Feasibility evolution on challenging constrained optimal control problems computed by \texttt{OdynSQP}.}
\label{fig:feasibility}
\end{figure}

We evaluated \texttt{OdynSQP} on a set of challenging trajectory optimization problems. 
Specifically, we considered three representative robotic systems with nonlinear dynamics subject to both inequality constraints (joint and control limits, friction-cone constraints) and equality constraints (end-effector targets and forward-dynamics consistency).
In each of the examples, we included state and control regularization terms.
Snapshots of the resulting optimal trajectories are reported in~\Cref{fig:odynsqp_to_results} and the feasibility evolution is depicted in~\Cref{fig:feasibility}. 

\texttt{OdynSQP} successfully optimized highly dynamic maneuvers, such as backflips on the G1 humanoid robot, while enforcing joint position, velocity, and torque limits together with contact wrench-cone constraints. 
It also enabled loco-manipulation, including the ``chicken-head'' task on the Unitree B1 quadruped equipped with the Z1 manipulator. 
In addition to joint and friction-cone constraints, we imposed bounding-box constraints on the end-effector, as illustrated in~\Cref{fig:odynsqp_to_results}. 
Furthermore, \texttt{OdynSQP} handled aerial manipulation scenarios requiring end-effector target constraints and thruster limits. From~\Cref{fig:feasibility}, we observe that \texttt{OdynSQP} decreased the constraint feasibility to high accuracy $(\textrm{i.e., }10^{-9})$ while allowing temporary increases in feasibility through its non-monotone line-search  globalization strategy exhibiting strong convergence capabilities.  
These experiments demonstrated the ability of \texttt{OdynSQP} to efficiently solve highly nonlinear optimal control problems across diverse robotic platforms.

\subsection{Model predictive control via \texttt{OdynSQP}}

\begin{figure*}[!t]
\centering
\href{https://www.youtube.com/watch?v=fJqjgdT2De0}{
    \includegraphics[width=\textwidth]{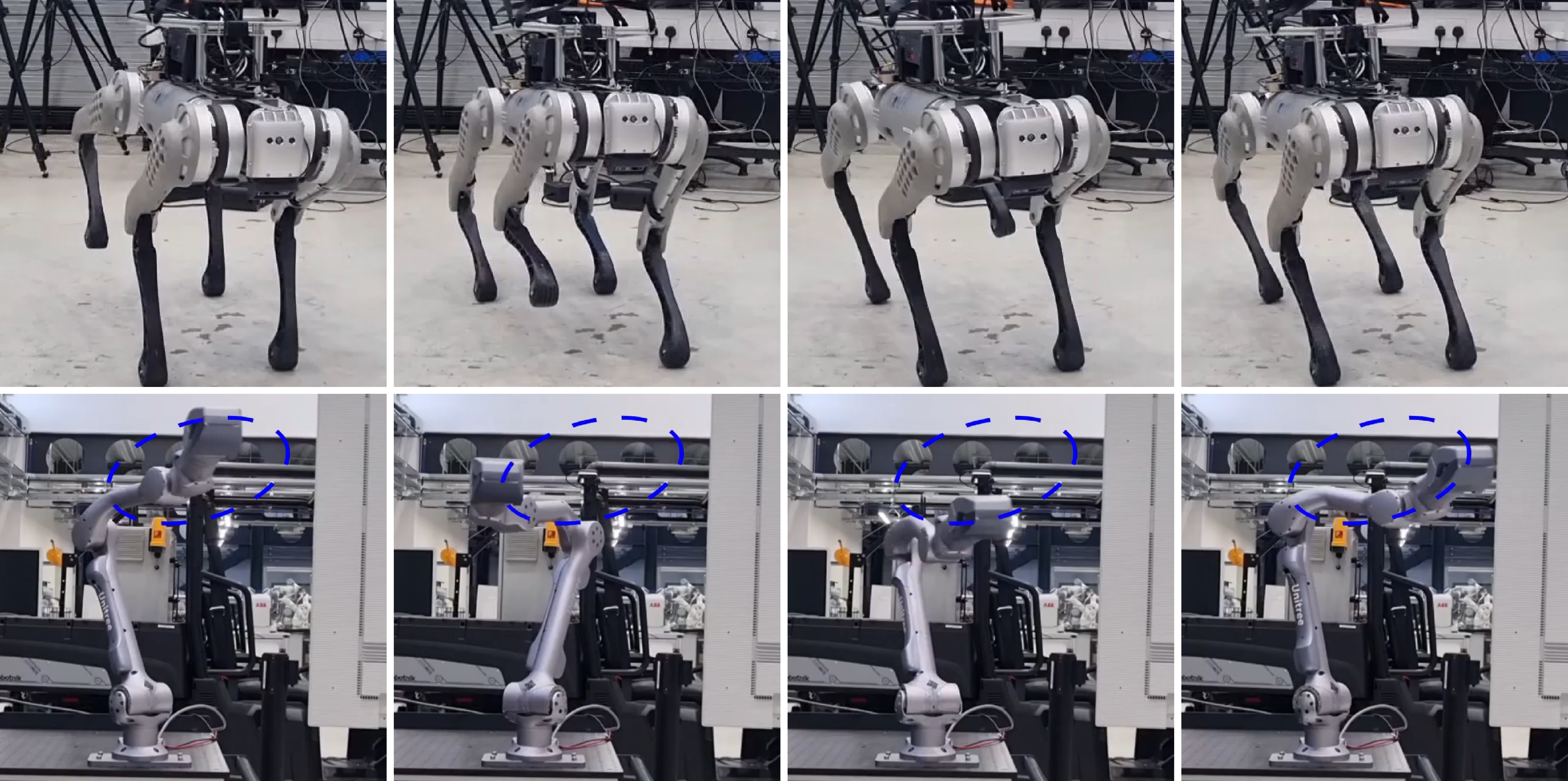}
}
\caption{Snapshots of~\gls{mpc} trajectories generated by \texttt{OdynSQP}. 
  (Top) B1 quadruped executing a quasi-static walking-in-place motion under contact and friction-cone constraints, while enforcing state and input constraints. 
  (Bottom) Z1 manipulator tracking a three-dimensional ellipsoidal end-effector trajectory (blue dashed curve) under state and input constraints. 
  Despite changing active sets and high-frequency updates, \texttt{OdynSQP} achieved accurate tracking and stable behavior across successive~\gls{mpc} iterations.To watch the video, click the picture or see \href{https://www.youtube.com/watch?v=fJqjgdT2De0}{https://www.youtube.com/watch?v=fJqjgdT2De0}.}
  \label{fig:mpc_results}
\end{figure*}

We evaluated \texttt{OdynSQP} in a model predictive control setting on two Unitree robotic platforms: the B1 quadruped and the Z1 manipulator. These experiments assess the tracking performance and real-time capabilities of \texttt{OdynSQP} in MPC loops (see~\Cref{fig:mpc_results}).

\begin{figure}[!t]
  \centering
  \includegraphics[width=\linewidth]{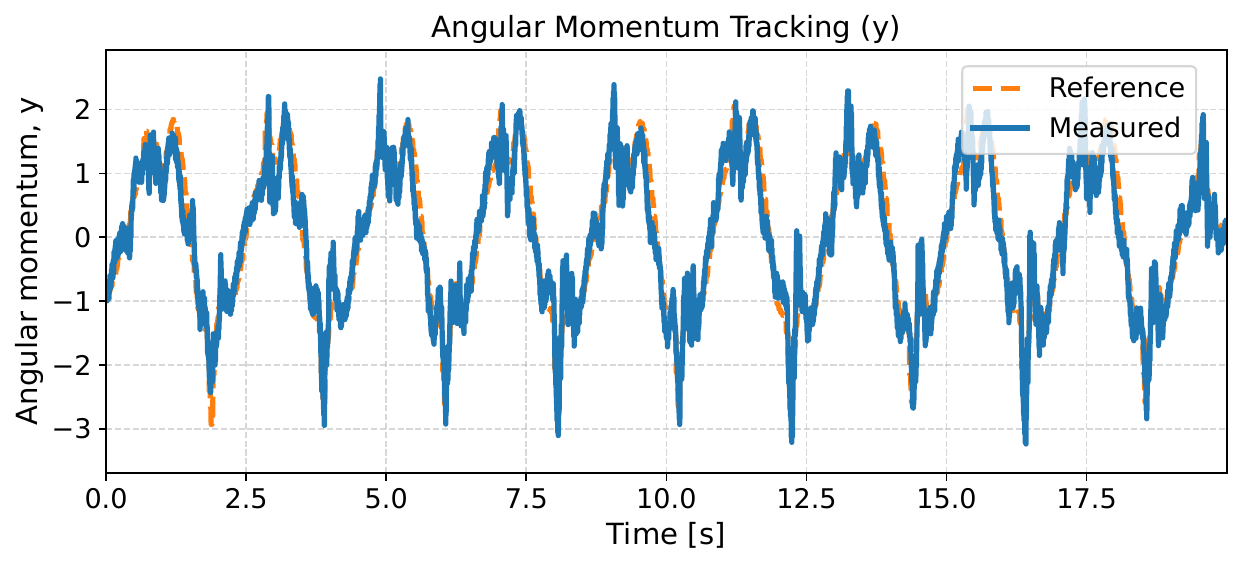}
  \par\medskip
  \includegraphics[width=\linewidth]{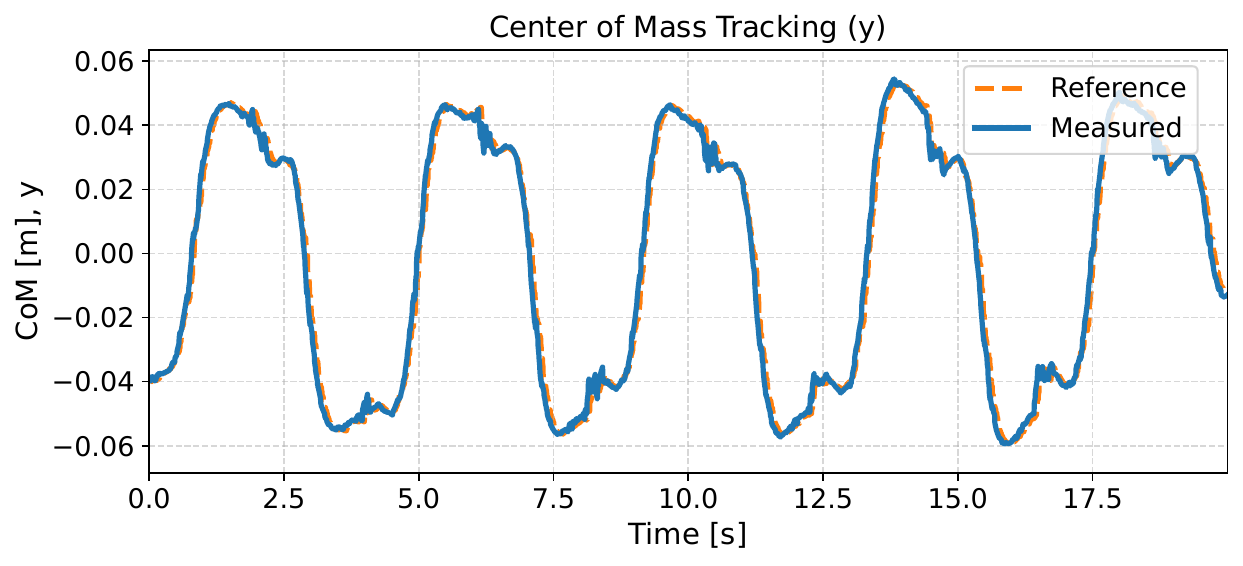}

  \caption{Model predictive control tracking performance on the Unitree B1 quadruped using \texttt{OdynSQP}. The plots show the tracking of the center-of-mass position (bottom) and angular momentum (top) along the $y$-axis during a quasi-static walking-in-place motion.}
  \label{fig:b1_mpc_tracking}
\end{figure}

The first experiment considers the B1 quadruped performing a quasi-static walking-in-place motion under contact constraints.
Our~\gls{mpc} formulation enforced friction-cone constraints together with state and input constraints, while tracking a desired center-of-mass and angular momentum trajectory, as shown in~\Cref{fig:b1_mpc_tracking}.
The~\gls{mpc} horizon was defined with nine nodes discretized at \(33\,\mathrm{ms}\), corresponding to a prediction horizon of \(0.3\,\mathrm{s}\), and the control loop runs at \(30\,\mathrm{Hz}\). Despite frequent changes in the active set induced by contact transitions, \texttt{OdynSQP} maintained stable convergence across successive~\gls{mpc} iterations, resulting in smooth and consistent motion generation.

\begin{figure}[!t]
  \centering
  \includegraphics[width=\linewidth]{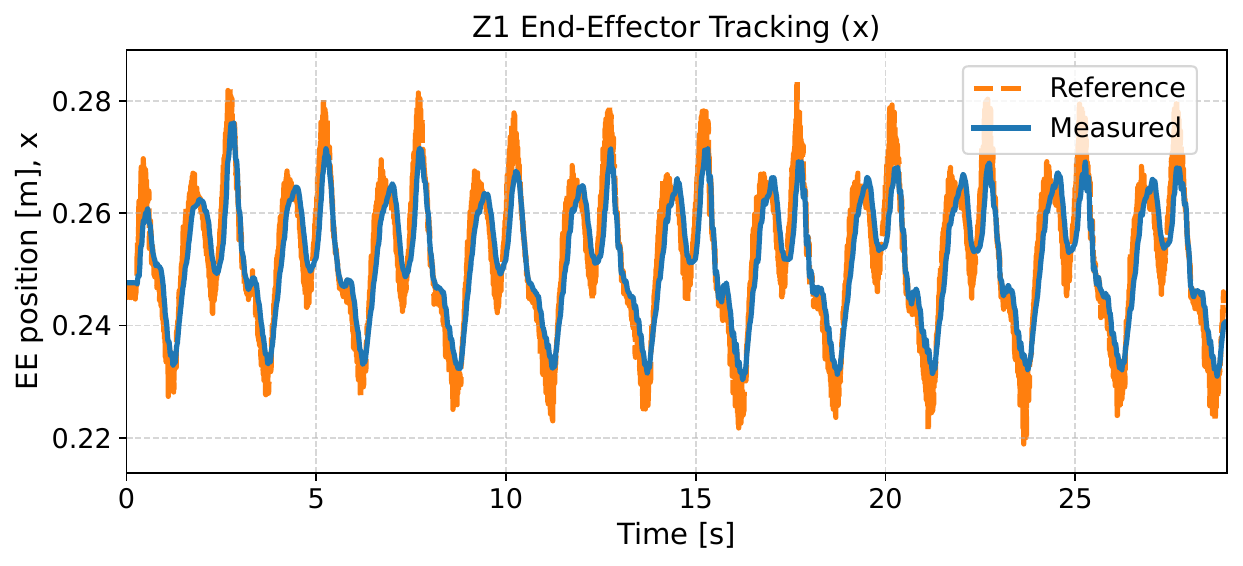}
  \par\medskip
  \includegraphics[width=\linewidth]{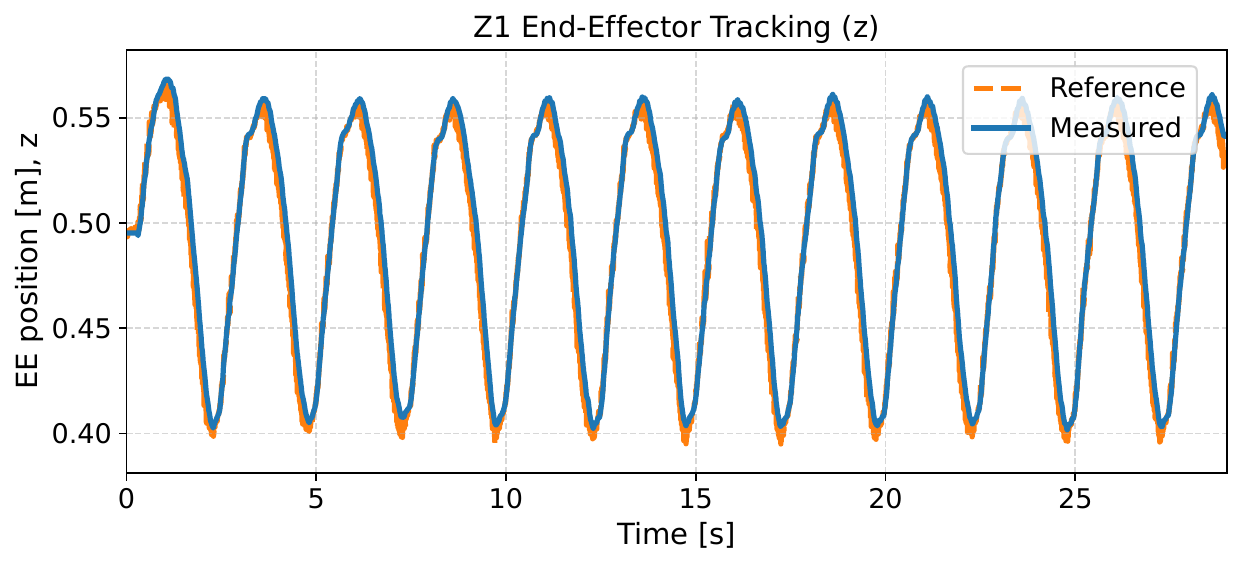}

  \caption{Model predictive control tracking performance on the Unitree Z1 manipulator using \texttt{OdynSQP}. The plots show the tracking of the end-effector position along selected axes while following a three-dimensional ellipsoidal trajectory.}
  \label{fig:z1_mpc_tracking}
\end{figure}

Solver statistics further highlight the efficiency of \texttt{OdynSQP}.
For the Z1 manipulator, the average solve time was \(13.5\,\mathrm{ms}\), with a maximum of \(19.6\,\mathrm{ms}\), remaining strictly below the \(20\,\mathrm{ms}\) control period.
For the B1 quadruped, the average solve time was \(19.7\,\mathrm{ms}\), with a maximum of \(29.4\,\mathrm{ms}\), also below the \(33.3\,\mathrm{ms}\) control period.
In both cases, the solver was limited to a single~\gls{sqp} iteration per~\gls{mpc} step, demonstrating the capability of \texttt{OdynSQP} to operate reliably under real-time constraints.

These results demonstrated that \texttt{OdynSQP} operates reliably across different~\gls{mpc} regimes, ranging from locomotion to manipulation, and provides an efficient backend for solving sequences of structured~\glspl{qp} in real-time robotics applications.

\subsection{Fast resolution of contact physics via \texttt{ODYNSim}}
\label{sec:odynsim_results}

We evaluated \textsc{Odyn}, \textsc{ProxQP}, and \textsc{OsQP} in the context of contact-physics simulation. 
The benchmark consists of a pushed-cube scenario in which contact forces are computed by solving a quadratic program at each simulation step.
This setup provides a controlled environment to compare solver robustness, convergence behavior, and computational efficiency. 

A single cube (edge length $0.20\,\mathrm{m}$, mass $15\,\mathrm{kg}$) rests on a fixed plane, an external horizontal force was applied throughout the simulation, causing it to slide across the surface. The force magnitude was set to $2 \mu m g \approx 147.15\,\mathrm{N}$, and its direction alternated every $0.5\,\mathrm{s}$ between the world $x$- and $y$-axes. 
Within each interval, the force was ramped linearly over the first $0.3\,\mathrm{s}$ and then held constant; the sign of the force was deterministically flipped at each interval. 

At every time step, the simulator enforced non-penetration and Coulomb friction constraints as formulated in~\Cref{eq:contact_qp}. 
Contacts were modeled using pyramidal friction cones with coefficient $\mu = 0.5$. 
The time step was set to $dt = 0.004\,\mathrm{s}$, and this \gls{qp} problem was solved at each step with $\epsilon_a=10^{-8}$ and $\epsilon_r = 0$.

\begin{figure}[!t]
  \centering

  \includegraphics[width=\linewidth]{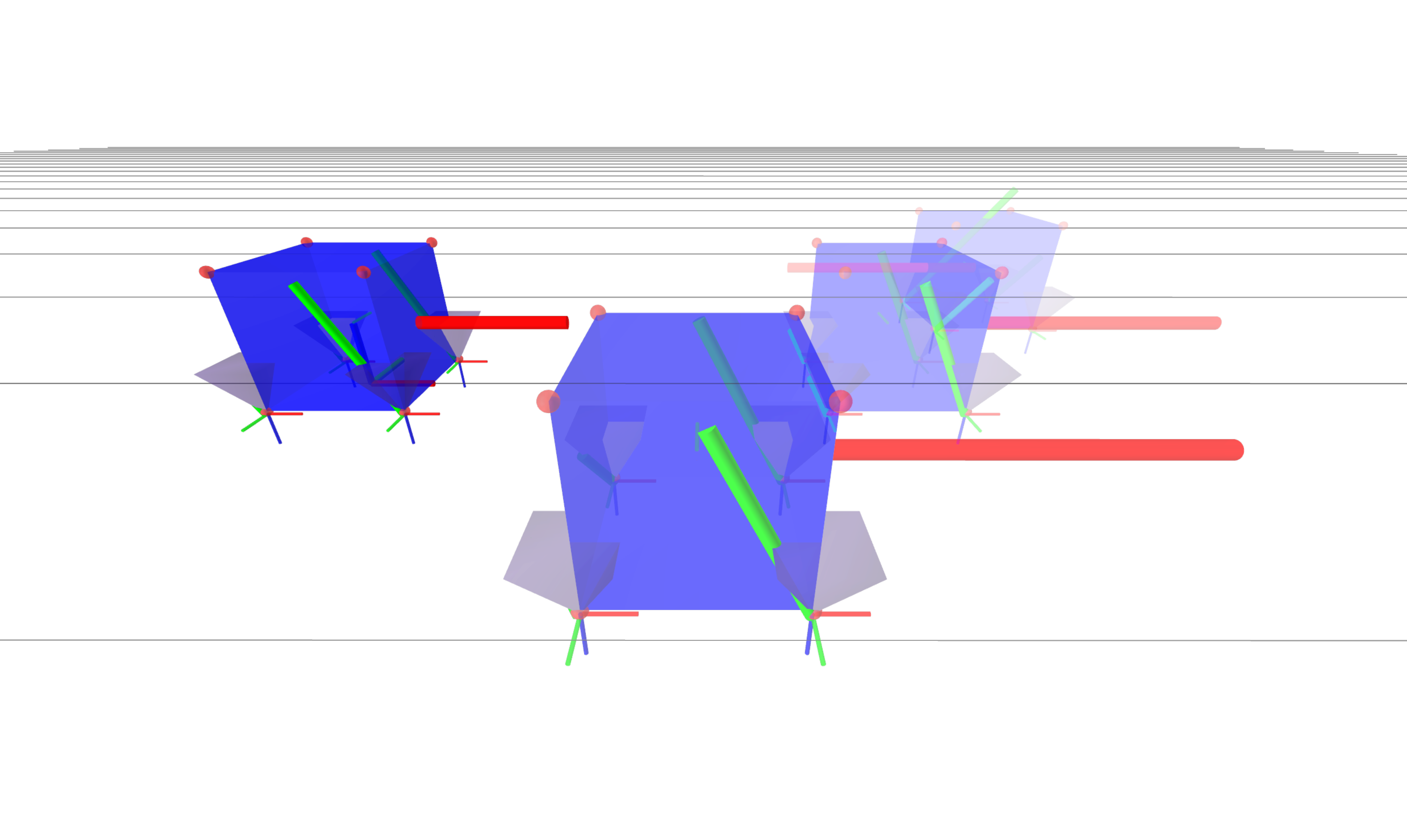}

  \par\medskip

  \includegraphics[width=\linewidth]{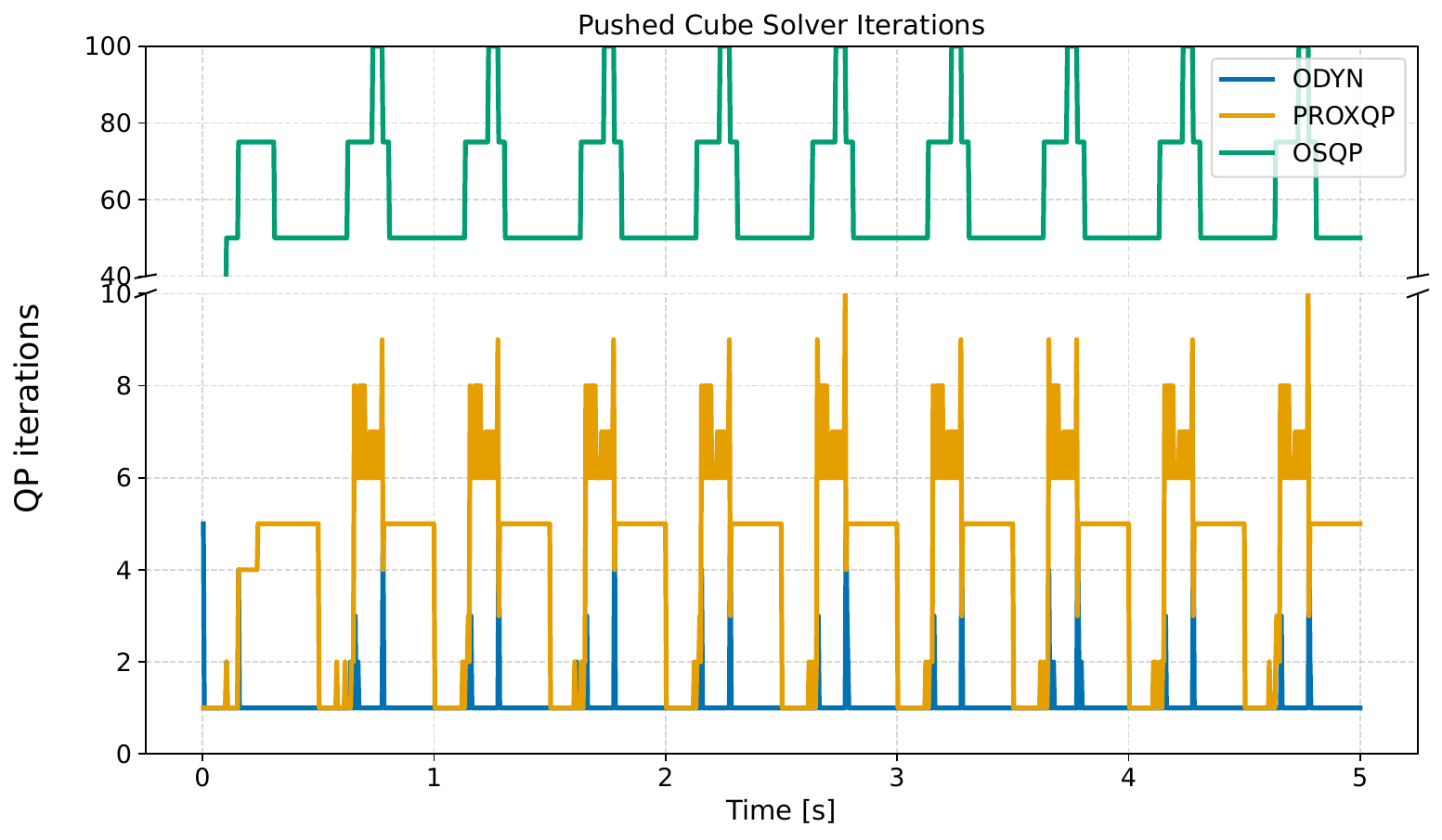}

  \caption{Pushed cube contact simulation resolved via~\gls{qp}-based contact dynamics. (Top) illustrate the cube's interaction with the plane under unilateral frictional contact constraints at different time instants.
  (Bottom) compares the number of solver iterations required per time step for \textsc{Odyn}, \textsc{ProxQP}, and \textsc{OsQP}.}
  \label{fig:odyn_sim}
\end{figure}

As shown in~\Cref{fig:odyn_sim}, \textsc{Odyn} consistently required fewer iterations per time step than \textsc{ProxQP} and \textsc{OsQP}. 
In particular, \textsc{OsQP} exhibited significantly higher iteration counts, reflecting the limited efficiency of first-order operator-splitting methods in exploiting warm starts at high accuracy.
This behavior is evident even when successive \glspl{qp} differ only marginally due to incremental state and contact updates.

When transitions occur between sticking and sliding contact, the optimal solution changes more substantially. 
Even in these cases, \textsc{Odyn} converged in fewer iterations than \textsc{ProxQP}. 
Overall, these results indicate that \textsc{Odyn} is well suited to real-time contact-dynamics simulation, where sequences of related \glspl{qp} must be solved efficiently.

\subsection{Learning to play Sudoku via \texttt{ODYNLayer}}\label{sec:odynlayer_results}
\texttt{ODYNLayer} remains differentiable, even with degenerate problems, thanks to its smoothed~\gls{ncp} regularization and proximal primal--dual structure.
This brings a more principled way to develop differentiable without relying on sub-gradients.
To demonstrate \texttt{ODYNLayer}'s capabilities, we trained a~\gls{qp} optimization to learn playing Sudoku as similarly done in~\cite{optnet}.

\begin{figure*}[t]
    \centering
        \begin{minipage}[t]{0.70\textwidth}
            \vspace{0pt}
            \centering
            \includegraphics[height=0.23\textheight, trim=60 0 85 0, clip]{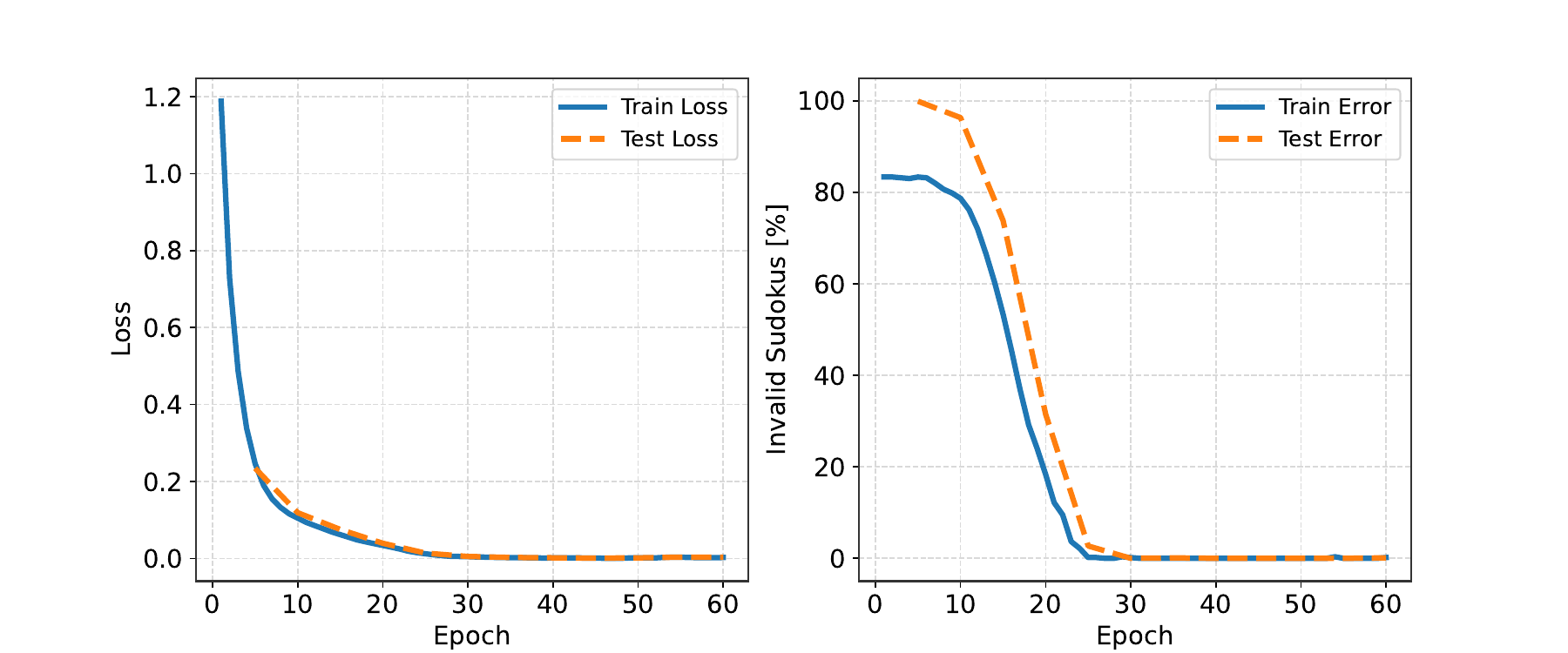}
            \par\footnotesize\textbf{(a)} Double Precision (\textit{float64})
            \includegraphics[height=0.23\textheight, trim=60 0 85 0, clip]{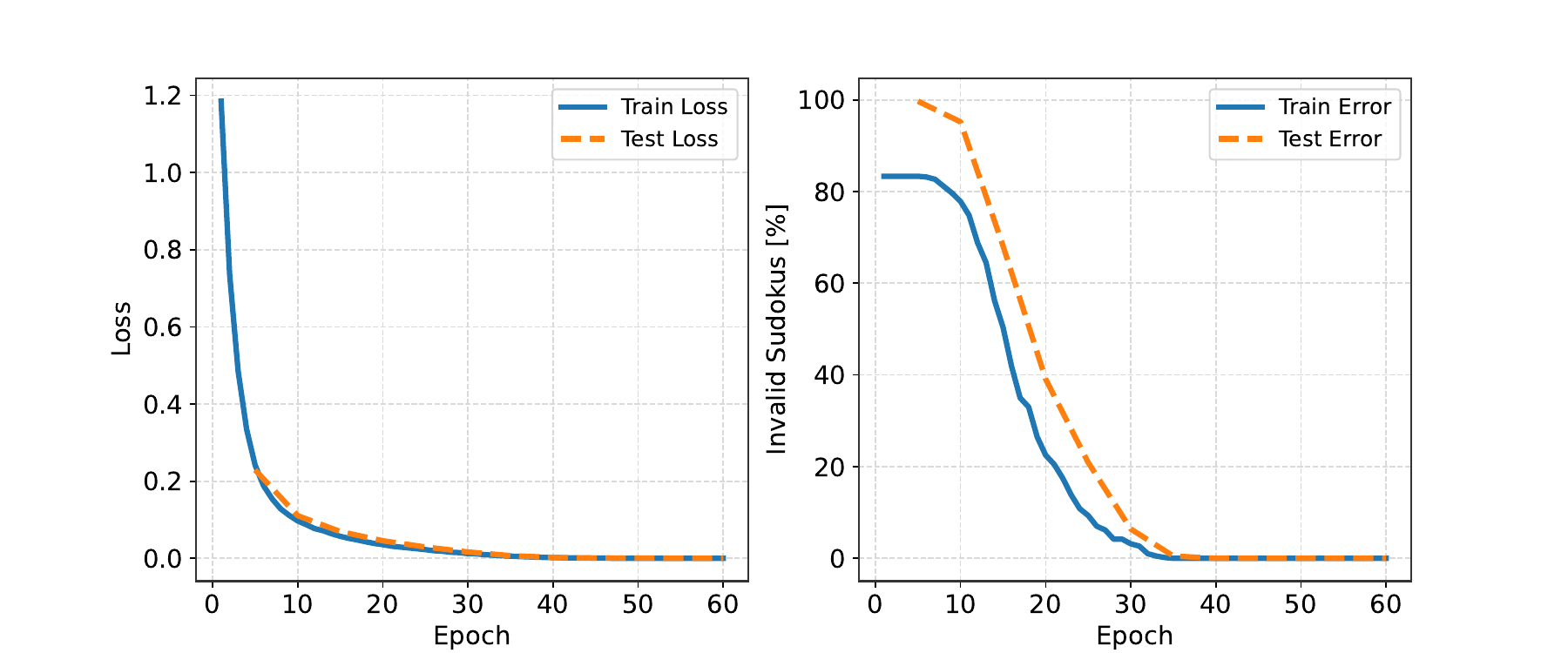}
            \par\footnotesize\textbf{(b)} Single Precision (\textit{float32})
        \end{minipage}%
        \hfill
        \begin{minipage}[t]{0.29\textwidth}
            \vspace{0pt}
            \centering

            \includegraphics[height=0.46\textheight,, trim=10 0 0 -20, clip]{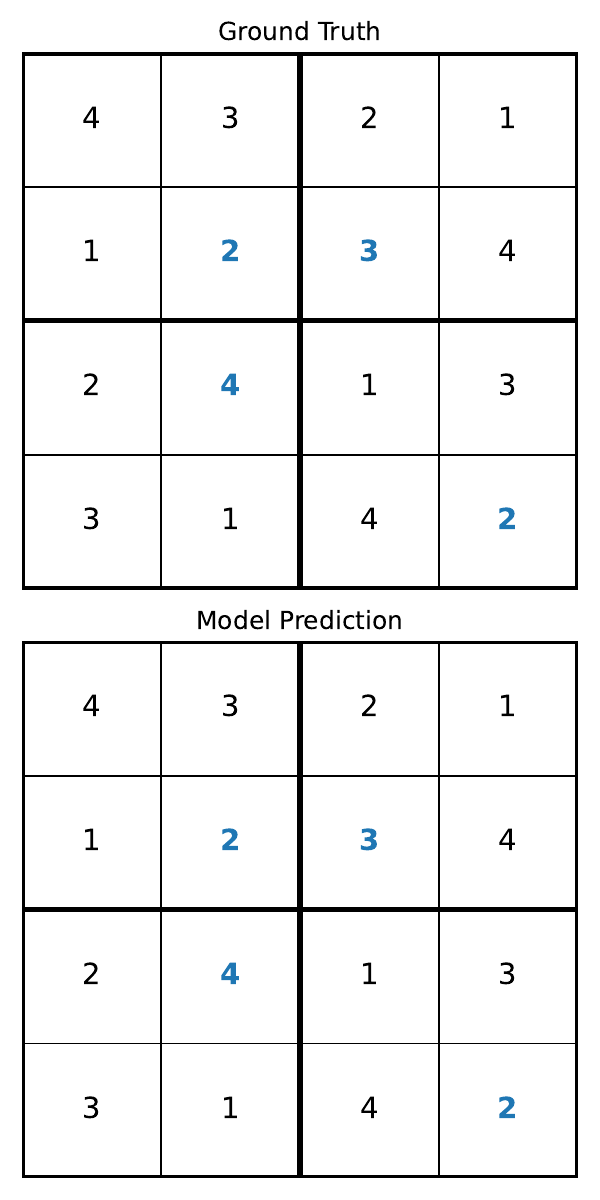}
            \par\footnotesize\textbf{(c)} Ground Truth vs.\ Model Prediction
        \end{minipage}%
    \caption{Effect of \texttt{ODYNLayer} precision on convergence and invalid-Sudoku rate.
    (a) With double precision (\textit{float64}), it exhibited stable convergence with a rapid decline in invalid Sudoku solutions.
    (b) With single precision (\textit{float32}), it showed slightly slower convergence in terms of invalid Sudoku count, a degradation that is attributed to \textsc{Odyn}'s single-precision convergence tolerance and randomized-parameter initialization.
    (c) Example reconstruction comparing ground truth with the model prediction; digits shown in blue denote the fixed givens supplied to \texttt{ODYNLayer}.
    These results highlight how numerical precision impacts both convergence speed and true Sudoku constraint satisfaction (invalid Sudoku count).
    True Sudoku constraint satisfaction implies that the learned linear constraint basis accurately captures the underlying Sudoku structure.}
    \label{fig:sudoku}
\end{figure*}

Sudoku is a logic-based puzzle played on a $N\times N$ grid divided into $n\times n$ subgrids (``blocks''), with $n=3$ and $N=n^2=9$ typically.
Sudoku is inherently a binary problem as it encodes digit assignments using binary variables $x_{i,j,k}\in\{0,1\}$ indicating whether cell $(i,j)$ that is located inside a $n\times n$ sub-block~$b$, contains digit~$k$ for $i,j,k\in\{1,\ldots,N\}$.
Theses rules are captured by equality constraints enforcing that every row, column, and $n\times n$ sub-block contains each digit exactly once:
\begin{equation}
\begin{aligned}
&\sum_{i=1}^{N} x_{i,j,k} = 1, &&\forall i,k, &
&\sum_{j=1}^{N} x_{i,j,k} = 1, &&\forall j,k,\\
&\sum_{(i, j)\in b} x_{i,j,k} = 1, &&\forall b,k,
\end{aligned}
\label{eq:sudoku_constraints}
\end{equation}
with pre-filled cells (``givens'') imposed by fixing $x_{i,j,k_0}=1$.

\subsubsection{Formulation}
To integrate this structure into a differentiable pipeline, we relaxed the integer nature of the decision variables by considering bounded real numbers (i.e.. $0\le x_{i,j,k}\le 1$), transforming it into a~\gls{lp}.
This relaxation is exact, since the constraint matrix defined by~\Cref{eq:sudoku_constraints} is totally unimodular.
This in turn implies that all extreme points of the feasible polytope are integer-valued~\cite{AlexLP}.
Thus, solving this~\gls{lp} recovers valid Sudoku solutions without enforcing integrality explicitly.
Specifically, we proposed to formulate a strictly feasible, regularized relaxation of this~\gls{lp} via \texttt{ODYNLayer}, i.e.,
\begin{equation}\label{eq:feasible_sudoku_qp}
\begin{aligned}
\min\limits_{\mathbf x \in \mathbb{R}^{n^6}} \quad
    & \tfrac{\epsilon}{2}\,\|\mathbf x\|_2^2 + \mathbf c^\transpose \mathbf x \\
\text{subject to} \quad
    & \mathbf A(\boldsymbol \theta)\,\mathbf x = \mathbf b(\boldsymbol \theta), \\
    & \mathbf 0 \leq \mathbf x \leq \mathbf 1,
\end{aligned}
\end{equation}
where $\mathbf{x}$ encodes the relaxed digit-assignment variables of the Sudoku instance and $\epsilon = 0.1$ is a small Tikhonov regularization parameter~\cite{tikhonov1977} to follow \textsc{OptNet}'s problem setup~\cite{optnet}.
Givens are incorporated as soft constraints through a linear penalty vector $\mathbf{c} = -\mathbf{x}_{k_0}$, which biases the optimizer toward satisfying the given entries while preserving differentiability.
To enforce strict feasibility, we set $\mathbf{b}(\boldsymbol{\theta})$ to be in the range of $\mathbf A(\boldsymbol \theta)$.
This enabled smooth implicit differentiation through the~\gls{qp}, while maintaining the exact combinatorial structure of the Sudoku constraints.

\subsubsection{Training capabilities}
We evaluated \texttt{ODYNLayer} on a diverse collection of $4 \times 4$ Sudoku instances. 
For training, we generated $500$ randomly sampled Sudoku instances and reserved a disjoint set of $1000$ additional instances for testing. 
This setup yields a substantially smaller training set than that used by \textsc{OptNet} ($9000$ Sudoku instances).
However, we employed additional training epochs to ensure reliable convergence under this more limited data regime.

The results are summarized in~\Cref{fig:sudoku}.
Training proceeded by minimizing the discrepancy between the predicted digits $\hat{\mathbf x}_{i,j}(\boldsymbol \theta)$ and the ground-truth values $\mathbf x^*_{i,j}$ for each cell $(i,j)$ across batches of size $B$, namely,
\begin{equation}
\label{eq:sudoku_loss}
\mathcal{L}(\boldsymbol\theta)
=
\frac{1}{B N^2}
\sum_{b=1}^{B}
\sum_{i=1}^{N}
\sum_{j=1}^{N}
\bigl\|
  \hat{\mathbf x}^{(b)}_{i,j}(\boldsymbol\theta)
  - \mathbf x^{*(b)}_{i,j}
\bigr\|_2^2.
\end{equation}
This loss function served as a surrogate objective, guiding the identification of a linear-constraint basis that faithfully encodes the Sudoku rules defined in~\Cref{eq:sudoku_constraints}.

\subsubsection{Effect of multiple setups}
We analyzed the capabilities of \texttt{ODYNLayer} across different numerical configurations, considering both sparse and dense backends as well as full-\gls{kkt} and condensed-\gls{kkt} factorizations. 
The results of this ablation study are reported in~\Cref{tab:sudoku_ablation}. 
\texttt{ODYNLayer} produces zero invalid Sudoku predictions in all configurations except for the condensed-\gls{kkt} factorization in single precision. 
This behavior is consistent with the well-known numerical instability associated with condensation-based approaches.

Furthermore, for the full-\gls{kkt} factorization, the faster convergence rate observed with single precision is explained by the need to relax the solver stopping tolerances—set to $10^{-9}$ in double precision and $10^{-5}$ in single precision—when operating at reduced numerical accuracy.

\begin{table}[t]
\caption{Sudoku ablation study with \texttt{ODYNLayer}: We indicate the average solver iterations per epoch (iters) during training and the number of invalid Sudoku predictions (\#errors) in the out-of-distribution test-set ($1000$ Sudoku instances).
The full-\gls{kkt} factorization setup is indicated by ``Full'' and the condensed by ``Cond.''.}
\label{tab:sudoku_ablation}
\centering
\setlength{\tabcolsep}{3pt}
\renewcommand{\arraystretch}{0.9}
\small
\begin{tabular}{@{}llcccc@{}}
\toprule
\multicolumn{2}{l}{} &
\multicolumn{2}{c}{Double (\textit{float64})} &
\multicolumn{2}{c}{Float (\textit{float32})} \\
\cmidrule(r){3-4}\cmidrule(l){5-6}
 &  & \#errors & iters ($\pm \sigma$) & \#errors & iters ($\pm \sigma$) \\
\midrule
\multirow{2}{*}{Full} 
  & Dense  & 0  & $11.38\;(\pm 0.31)$  &  0 & $8.91\;(\pm 0.30)$  \\
  & Sparse & 0  & $11.16\;(\pm 0.29)$ & 0 & $8.98\;(\pm 0.30)$ \\
\midrule
\multirow{2}{*}{Cond.} 
  & Dense  & 0  & $11.36 \;(\pm 0.28)$  & - \textit{(max iters)}   & - \textit{(max iters)}  \\
  & Sparse & 0  & $11.14\;(\pm 0.28)$ & - \textit{(max iters)} & - \textit{(max iters)} \\
\bottomrule
\end{tabular}
\end{table}

\section{Conclusion}

This paper presented \textsc{Odyn}, a novel all-shifted non-interior-point~\gls{qp} solver based on shifted~\gls{ncp} functions and proximal primal--dual regularization. By relaxing strict interior-feasibility requirements while preserving path-following behavior, \textsc{Odyn} provides a principled alternative to classical interior-point methods. The resulting framework enables robust treatment of degenerate, ill-conditioned, and warm-started problems that frequently arise in robotics and AI workloads.

Extensive numerical experiments on standard benchmarks demonstrate that \textsc{Odyn} achieves competitive convergence performance on par with modern interior-point methods while exhibiting superior warm-start capabilities compared to state-of-the-art augmented Lagrangian and operator-splitting approaches. Beyond standalone~\gls{qp} benchmarks, \textsc{Odyn} was validated in representative robotic and AI applications, including sequential quadratic programming, contact-dynamics simulation, and differentiable optimization layers, illustrating its versatility across robotics, physics simulation, and learning pipelines.

In summary, \textsc{Odyn} blends the advantages of primal--dual interior-point methods and generalized augmented Lagrangian techniques, offering a powerful and flexible foundation for the development of the next generation of optimization solvers for robotics, AI, and numerical optimization.


\appendix
\section*{A. All-shifted complementarity constraints}
\label{app:appendix}

From the~\gls{kkt} conditions of the perturbed~\gls{ncp} Lagrangian (\Cref{eq:all_shifted_KKT_conditions}), namely 
$\nabla_{\vxi}\tilde{\mathcal{P}} = \vzero$ and $\nabla_{\vw}\tilde{\mathcal{P}} = \vzero$, we obtain
\begin{equation}\label{eq:all_shited_complementarity}
-\mu\vxi^{-1} + \vw + \frac{\rho_n}{2}(\vs - \sE) + \frac{\rho_n}{2}(\vw - \wE) = \vzero.
\end{equation}

The condition $\vncpFun{}(\vs, \vw;\mu) = \vzero$ implies $\vs = \vxi$. Substituting into~\Cref{eq:all_shited_complementarity} yields the all-shifted complementary constraints:
\begin{equation}
\vs \circ \vw = \mu\bone + \frac{\rho_n}{2}\,\vs \circ (\sE - \vs + \wE - \vw).
\end{equation}

Finally, non-negativity of $\vs$ and $\vw$ follows directly from $\vncpFun{}(\vs, \vw;\mu) = \vzero$.

\bibliography{references}



\end{document}

%% file: math_commands.tex
\usepackage{amssymb}

\newcommand{\vct}[1]{\mathbf{#1}}         
\newcommand{\mtr}[1]{\mathbf{#1}}         
\newcommand{\bzero}{\mathbf{0}}           
\newcommand{\vzero}{\mathbf{0}}           
\newcommand{\bone}{\mathbf{1}}
\newcommand{\transpose}{\intercal}

\newcommand{\norm}[1]{\left\lVert #1 \right\rVert}

\DeclareMathOperator{\diag}{diag}

\newcommand{\calN}{\mathcal{N}}
\newcommand{\calP}{\mathcal{P}}

\newcommand{\calL}{\mathcal{L}}

\newcommand{\R}{\mathbb{R}}

\newcommand{\matQ}{\mtr{Q}}
\newcommand{\vecC}{\vct{c}}
\newcommand{\matA}{\mtr{A}}
\newcommand{\vecB}{\vct{b}}
\newcommand{\matG}{\mtr{G}}
\newcommand{\vecH}{\vct{h}}

\newcommand{\vx}{\vct{x}}
\newcommand{\vs}{\vct{s}}
\newcommand{\vxi}{\boldsymbol{\xi}}

\newcommand{\vy}{\vct{y}}
\newcommand{\vz}{\vct{z}}
\newcommand{\vw}{\vct{w}}

\newcommand{\vq}{\vct{q}}

\newcommand{\U}{\vct{U}}          
\newcommand{\Ut}{\tilde{\U}}      
\newcommand{\vlam}{\boldsymbol{\lambda}}

\newcommand{\matJ}{\mtr{J}}        

\newcommand{\vecM}{\vct{m}}
\newcommand{\vecZeta}{\boldsymbol{\zeta}}
\newcommand{\vecEta}{\boldsymbol{\eta}}

\newcommand{\vr}{\vct{r}}

\newcommand{\vtres}[1]{\tilde{\vct{r}}_{#1}}
\newcommand{\vtresd}{\vtres{d}}

\newcommand{\vtresp}{\vtres{p}}

\newcommand{\NatMerit}{\mathcal{M}}

\newcommand{\NipNbr}[1]{\mathcal{N}\!\left(#1\right)}

\newcommand{\vncpFun}[1]{\boldsymbol{\phi}^{#1}}

\newcommand{\ncpJac}[1]{\boldsymbol{\Phi}_{#1}}
\newcommand{\pertncpJac}[1]{\tilde{\boldsymbol{\Phi}}_{#1}}

\newcommand{\rhod}{\rho_{d}}      
\newcommand{\rhoe}{\rho_{e}}      
\newcommand{\rhoi}{\rho_{i}}      
\newcommand{\rhon}{\rho_{n}}      
\newcommand{\rhoBar}{\bar{\rho}}  
\newcommand{\vRhoBar}{\boldsymbol{\rhoBar}} 

\newcommand{\xE}{\vx_{E}}
\newcommand{\yE}{\vy_{E}}
\newcommand{\zE}{\vz_{E}}
\newcommand{\sE}{\vs_{E}}
\newcommand{\wE}{\vw_{E}}

\newcommand{\Qtil}{\tilde{\matQ}}
\newcommand{\matP}{\mtr{P}}
\newcommand{\matD}{\mtr{D}}
\newcommand{\matH}{\mtr{H}}       

\newcommand{\matS}{\mtr{S}}
\newcommand{\matZ}{\mtr{Z}}
\newcommand{\Id}{\mtr{I}}

\newcommand{\dx}{\Delta\vx}
\newcommand{\dy}{\Delta\vy}
\newcommand{\dz}{\Delta\vz}
\newcommand{\ds}{\Delta\vs}
\newcommand{\dU}{\Delta \U}
\newcommand{\dl}{\Delta\vlam}

\newcommand{\alphals}{\alpha}
\newcommand{\gammaLS}{\gamma}
\newcommand{\etaLS}{\eta}

\newcommand{\nq}{{n_{q}}}
\newcommand{\nx}{{n_{x}}}
\newcommand{\ntau}{{n_{u}}}
\newcommand{\nc}{{n_{c}}}
\newcommand{\state}{\vct{x}}
\newcommand{\control}{\vct{u}}
\newcommand{\force}{\boldsymbol{\lambda}}
\newcommand{\delassusMat}{\boldsymbol{\Lambda}}
\newcommand{\deSaxce}{\boldsymbol{\Gamma}}
\newcommand{\freeContVel}{\boldsymbol{\sigma}}
\newcommand{\contactJac}{\mathbf{J}_c}
\newcommand{\massMat}{\mathbf{M}}
\newcommand{\xu}{\mathbf{w}}
\newcommand{\pos}{\vct{q}}
\newcommand{\vel}{\vct{v}}
\newcommand{\dynFun}{\mathbf{f}}
\newcommand{\eqFun}{\mathbf{h}}
\newcommand{\ineqFun}{\mathbf{g}}
\newcommand{\stateManif}{\mathcal{X}} 
